\documentclass[letterpaper,journal]{IEEEtran}

\usepackage{amsmath,amsfonts}
\usepackage{array}
\usepackage[caption=false,font=normalsize,labelfont=sf,textfont=sf]{subfig}
\usepackage{textcomp}
\usepackage{stfloats}
\usepackage{float}
\usepackage{url}
\usepackage{verbatim}
\usepackage{graphicx}
\usepackage{times}
\usepackage{epsfig}
\usepackage{amssymb}
\usepackage{epstopdf}
\usepackage{makecell}
\usepackage{arydshln}
\usepackage{color}
\usepackage{multirow}
\usepackage{setspace}
\usepackage{booktabs}
\usepackage{enumerate}
\usepackage{pifont}
\usepackage{ragged2e}
\usepackage{tikz}

\newcommand{\revised}[1]{\textcolor{black}{#1}}

\newcommand{\myeqref}[1]{(\ref{#1})}
\newcommand{\ie}{\emph{i.e.}{}}
\newcommand{\eg}{\emph{e.g.}{}}
\newcommand{\etc}{\emph{etc}{}}
\newcommand{\circled}[1]{%
  \tikz[baseline=(char.base)]{
    \node[shape=circle,draw,inner sep=0pt,outer sep=0pt,minimum size=6pt,font=\tiny] (char) {#1};
  }%
}
\ifCLASSOPTIONcompsoc
  \usepackage[nocompress]{cite}
\else
  \usepackage{cite}
\fi
\usepackage{hyperref}
\hypersetup{colorlinks,%
	linkcolor=red,%
	citecolor=blue}

\begin{document}

\title{COVTrack++: Learning Open-Vocabulary Multi-Object Tracking from Continuous Videos via a Synergistic Paradigm} 

\author{Zekun Qian, Wei~Feng,~\IEEEmembership{Member,~IEEE}, 
Ruize~Han,~\IEEEmembership{Member,~IEEE},
Junhui~Hou,~\IEEEmembership{Senior~Member,~IEEE}

\thanks{Z. Qian is with the School of Computer Science and Technology, Tianjin University, Tianjin, China, and is also with the Department of Computer Science, City University of Hong Kong, Hong Kong SAR, China.}
\thanks{W. Feng is with the School of Computer Science and Technology, Tianjin University, Tianjin, China.}
\thanks{J. Hou is with the Department of Computer Science, City University of Hong Kong, Hong Kong SAR, China. Email: jh.hou@cityu.edu.hk}
\thanks{R. Han is with the Faculty of Computer Science and Artificial Intelligence, Shenzhen University of Advanced Technology.\\
(Z. Qian and W. Feng contributed equally to this paper).  (Corresponding authors: R. Han and J. Hou.)}
}


\maketitle

\begin{abstract}
\justifying
Multi-object tracking (MOT) has traditionally focused on a few specific categories, thereby restricting its applicability to real-world scenarios involving diverse objects.
Open-vocabulary multi-object tracking (OVMOT) addresses this limitation by enabling tracking of arbitrary categories, including novel objects unseen during training. 
However, current progress is constrained by two critical challenges: the lack of continuously annotated video data for model training, and the lack of a customized OVMOT framework to synergistically handle the sub-tasks, i.e., detection (including localization and classification) and association.
We address the data bottleneck by constructing C-TAO, the first continuously annotated training set for OVMOT, which increases annotation density by 26$\times$ over the original TAO and captures smooth motion dynamics and intermediate object states. 
For the framework bottleneck, we propose COVTrack++, a synergistic framework that achieves a bidirectional reciprocal mechanism between detection and association. 
This is realized across three modules: (1) Multi-cue adaptive fusion module dynamically balances the appearance, motion, and semantic cues for association feature learning; 
(2) Multi-granularity hierarchical aggregation module further exploits hierarchical spatial relationships in dense detections, where visible child nodes (e.g., object parts) assist occluded parent objects (e.g., whole body) for association feature enhancement; and
(3) Temporal confidence propagation module recovers flickering detections through high-confidence tracked objects, boosting low-confidence candidates across frames and creating a chain reaction that stabilizes trajectories. 
Extensive experiments on TAO demonstrate the state-of-the-art performance of COVTrack++, with novel TETA reaching 35.4\% and 30.5\% on validation and test sets, respectively, improving novel AssocA by 4.8\% and novel LocA by 5.8\% over previous methods, and show strong zero-shot generalization on BDD100K.
These results validate the effectiveness of the C-TAO dataset and the robustness of COVTrack++. 
\end{abstract}


\vspace{-5pt}
\section{Introduction}
\label{sec:intro}

\IEEEPARstart{M}{ulti}-object tracking (MOT) has traditionally been studied under the closed-set paradigm, where systems track a limited number of predefined object categories in controlled scenarios~\cite{dendorfer2020mot20, sun2022dancetrack, geiger2013vision}.
However, real-world applications such as Internet video understanding, autonomous driving, and robotic perception demand the ability to track arbitrary objects whose categories cannot be trained in advance.
This motivates \textit{open-vocabulary multi-object tracking} (OVMOT), which aims to localize, classify, and track diverse object categories in unconstrained videos, including both base classes seen during training and novel classes never encountered before~\cite{li2023ovtrack}. 

\begin{figure*}[!t]
    \centering
    \includegraphics[width=1.0\textwidth]{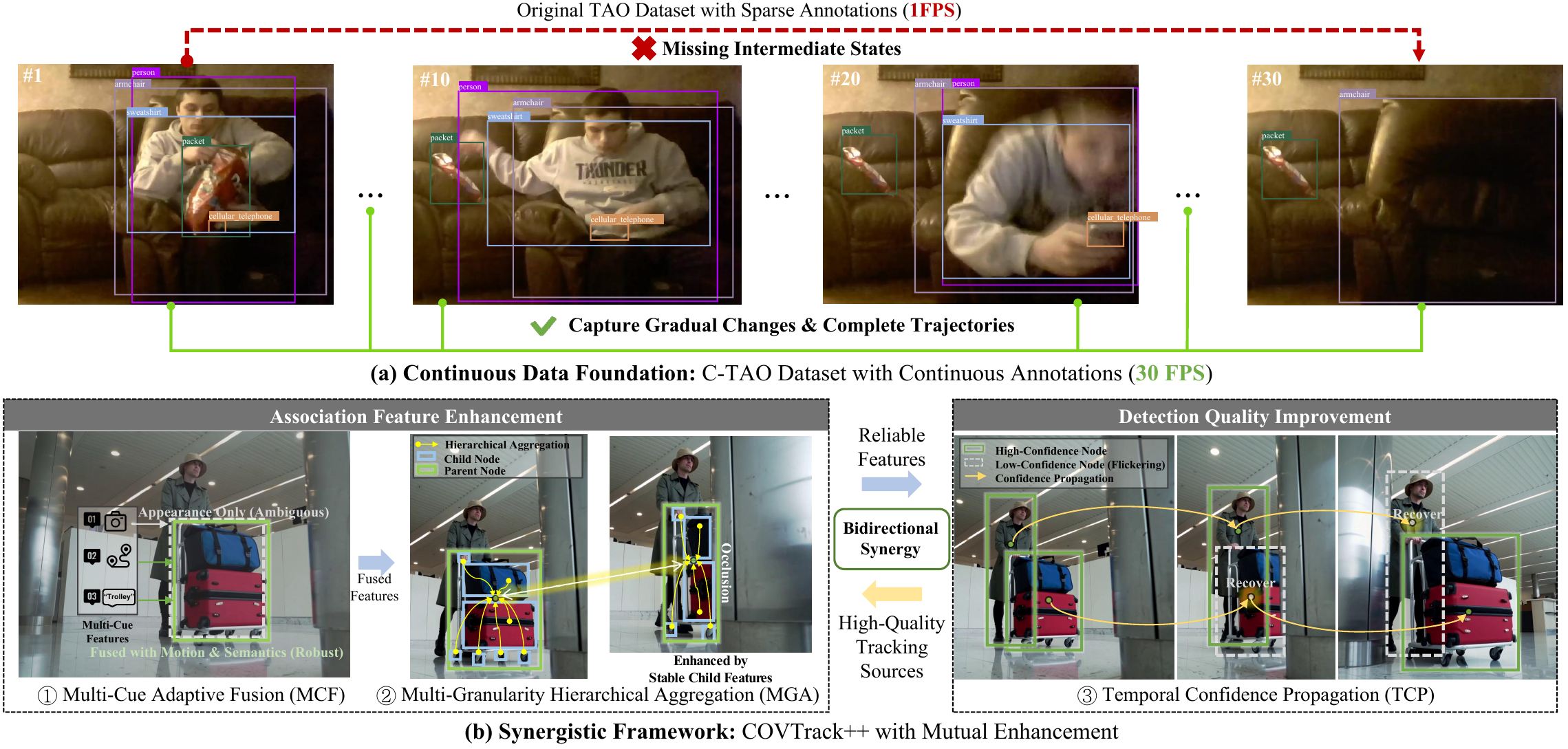}
    \caption{\revised{Overview of OVMOT challenges and our solutions. (a) Continuous data foundation: TAO's sparse annotations (every 30 frames) miss critical intermediate states. C-TAO provides continuous frame-by-frame annotations, enabling smooth trajectory learning. (b) Synergistic framework: COVTrack++ operates through two complementary stages that mutually reinforce each other. The association feature enhancement stage constructs robust association features through Multi-cue adaptive fusion (MCF) and Multi-granularity hierarchical aggregation (MGA). The detection quality improvement stage recovers flickering detections via Temporal confidence propagation (TCP).} 
    The proposed framework leverages reliable temporal propagation features from the association stage to improve the detection results; the recovered detections with boosted confidence serve as high-quality tracking sources for association in subsequent frames. This creates a bidirectional reciprocal mechanism that progressively improves both tracking continuity and detection quality.}
    \label{fig:overview}
\end{figure*}

\revised{Unlike closed-set MOT that detects and tracks a few specific types of targets, OVMOT is required to handle dense detections spanning hundreds of categories, including three sub-tasks: localization, obtaining the bounding boxes of category-agnostic objects; association, assigning the same tracking ID to the same object throughout the video; and classification, determining the category of each detected object.}
\revised{Recently, several methods~\cite{li2023ovtrack, Li_2024_CVPR, li2024slack, li2025open, li2025attention, li2025ovtr} have been proposed to handle the OVMOT problem. However, two main issues still exist: 1) data bottleneck, where OVMOT lacks training data with both diverse categories and temporally continuous annotations; and 2) framework bottleneck, where OVMOT lacks a dedicated framework that jointly handles localization, association, and classification.}

\revised{Specifically, for the data bottleneck, effective OVMOT learning requires training datasets that simultaneously satisfy two critical conditions: diverse object categories with varied motion patterns to enable open-vocabulary generalization, and dense temporal annotations to capture continuous motion dynamics and intermediate object states.}
However, existing datasets present a fundamental paradox.
On one hand, large-scale image datasets such as LVIS~\cite{gupta2019lvis} and SA-1B~\cite{kirillov2023segment} provide rich category diversity but completely lack temporal structure.
Current OVMOT methods~\cite{li2023ovtrack, Li_2024_CVPR, li2025open, li2025attention} compensate by generating synthetic image pairs to simulate consecutive frames, yet these pairs cannot capture realistic motion continuity, appearance transitions, or intermediate states such as occlusion and blur.
On the other hand, video tracking datasets like MOT20~\cite{dendorfer2020mot20} and DanceTrack~\cite{sun2022dancetrack} offer temporal information but are confined to a handful of categories (humans or vehicles), rendering them unsuitable for open-vocabulary scenarios.

For the framework bottleneck, different from classical MOT, the main challenges of OVMOT lie in simultaneously handling three tasks: localization, association and classification. 
Note that, beyond the additional requirement for classification, the localization and association in OVMOT also differ from those in classical MOT.
As illustrated in Fig.~\ref{fig:overview}(a), a person may be detected not only as a complete target but also as constituent parts such as sweatshirt, packet, and cellular telephone. This dramatic increase in detection density and category diversity makes OVMOT substantially more challenging than closed-set MOT, which also creates ambiguity in what to track and how to maintain identity consistency under occlusions or viewpoint changes.
However, most previous works for OVMOT concentrate on solving a single problem~\cite{li2024slack, li2025attention}, or designing the general MOT methods not specifically for the pain points in OVMOT~\cite{li2023ovtrack,Li_2024_CVPR, li2025open, li2025ovtr}.

In this work, we aim to address the above two challenges and answer the following questions: 
\revised{1) Is a temporally continuous OVMOT training dataset required to effectively benefit model training?}
\revised{2) How to build a customized framework for OVMOT that synergistically handles the sub-tasks of detection (including localization and classification) and association?}

\revised{First, to address the data bottleneck, one dataset satisfying the criteria for both diverse categories and continuous video sequences is TAO~\cite{dave2020tao}, a dataset containing 2,907 videos covering 833 diverse categories.}
However, TAO suffers from severe \textit{temporal discontinuity} due to its sparse annotation strategy (only every 30 frames, i.e., 1 fps).
As illustrated in Fig.~\ref{fig:overview}(a), the large temporal gaps between annotated frames (Frame 1 and Frame 30) create critical ambiguities: when the person and associated objects undergo significant state changes or disappear entirely between consecutive annotated frames, the intermediate process remains completely unobserved.
This temporal discontinuity manifests in multiple failure modes, including unobserved intermediate occlusions, abrupt viewpoint shifts, and drastic appearance variations (detailed in Section~\ref{sec:dataset}).
These disconnected trajectories force models to rely on heuristic interpolation or pseudo-labeling~\cite{li2024slack}, which lack temporal consistency and fail to capture the motion dynamics essential for tracking.

In this work, we construct \textit{C-TAO (Continuous TAO)}, which restores temporal continuity by providing frame-by-frame annotations for all videos in the TAO training set.
As shown in Fig.~\ref{fig:overview}(a), by manually labeling bounding boxes, identities, and category labels on previously unannotated frames, C-TAO captures gradual changes and complete trajectories, enabling models to observe smooth motion patterns, gradual appearance changes, and intermediate states that were previously invisible.
This continuous supervision serves as the necessary foundation for robust OVMOT learning, allowing the effective extraction of \textit{motion cues} derived from inter-frame spatial consistency alongside diverse category information.
C-TAO increases annotation density by 26$\times$ compared to the original TAO, providing the first continuously-annotated training dataset for OVMOT.

\revised{Second, to address the framework bottleneck, as shown in Fig.~\ref{fig:overview}(b), we propose COVTrack++, a synergistic framework with multi-task unification.}
As discussed above, OVMOT brings new challenges for object localization and association due to the diverse object categories and nested box relationships. \textit{Challenges bring opportunities}.
\revised{The key insight of our method is to fully explore and leverage the coupling and dependence among the three sub-tasks, thereby developing a unified and synergistic framework for OVMOT.}

For this purpose, with respect to association, we propose multi-cue adaptive fusion (MCF) and multi-granularity hierarchical aggregation (MGA) modules for effective association feature learning.
Besides the appearance cue used in previous OVMOT methods, MCF further integrates the continuous motion cue (\textit{enabled by C-TAO's continuous supervision}), and the semantic cue (\textit{driven by the diverse object categories in OVMOT}).
Moreover, MGA \textit{leverages the characteristics of nested box relationships in OVMOT} to improve target feature representations by leveraging their spatially related objects (bounding boxes).
With respect to detection, we propose temporal confidence propagation (TCP) to recover missed bounding boxes generated by the respective detection in each frame. TCP \textit{leverages the association features} to build the detection box connections between adjacent frames, whose training is made possible by C-TAO's continuous annotation.

Specifically, MCF integrates three complementary cues: appearance, motion, and semantic features. One issue for multi-cue fusion is the varied reliability across scenarios.
Semantic features may be highly confident for base categories but very low for novel categories, while motion cues become unreliable under camera shake or rapid movement. 
To address this, we develop an adaptive fusion mechanism with dual-perspective confidence estimation.
This dynamic balancing mechanism down-weights unreliable cues while amplifying stable ones, transforming uncertain individual features into robust fused representations. 
Next, we find that in OVMOT the dense detections commonly introduce hierarchical parent-child structures. 
When a parent is partially occluded, its child parts may remain visible and stable.
Thus we introduce MGA to selectively aggregate reliable child evidence and strengthen parent representations.
Finally, an observation of the detection is that the confidence scores are sensitive to subtle appearance changes, i.e., a valid object can drop below the threshold in the next frame, causing flickering trajectories.
To address this, we propose TCP to recover suppressed detections by propagating confidence along reliable temporal correspondences.
Together, the above modules form a unified synergistic OVMOT pipeline in which stronger association features enable better detection recovery, which in turn provides higher-quality tracking sources. 




The main contributions of this work are as follows. 
First, we construct C-TAO, the first continuously-labeled training dataset for OVMOT, by providing frame-by-frame annotations for all TAO training videos and increasing annotation density by 26$\times$. \revised{We further add C-TAO-val to support fine-grained temporal-continuity evaluation.}
Second, we propose an MCF mechanism that dynamically balances the appearance, motion, and semantic cues through dual-perspective confidence estimation.
Third, we propose an  MGA strategy to exploit parent-child spatial relationships in dense detections, leveraging reliable child nodes to enhance their parent features.
Finally, we propose TCP to recover suppressed detections via bipartite graph matching across frames, forming a bidirectional reciprocal mechanism between detection and association in OVMOT.

{Extension from Conference Version.}
This paper is an extension of our prior work~\cite{qian2025covtrack}, namely COVTrack, published at ICCV 2025.
The major extensions are summarized as follows.
First, we propose a \revised{\textit{multi-granularity hierarchical aggregation mechanism}}, MGA, in Section~\ref{sec:MGA} to exploit hierarchical spatial relationships in the dense detection scenarios of OVMOT, which further \textit{enhances the association features generated by~\cite{qian2025covtrack}}, enabling robust tracking under occlusion and viewpoint changes.
Second, we propose a \revised{\textit{temporal confidence propagation mechanism}}, TCP, in Section~\ref{sec:tcp}, a graph-based framework that recovers low-confidence detections through class-conditional confidence propagation across frames, addressing the detection flickering problem; TCP leverages the video priors to \textit{directly improve the original frame-level detection} without replacing the detector.
\revised{We also add C-TAO-val as a continuously annotated validation split, enabling dense evaluation of trajectory continuity beyond the sparse TAO validation protocol.}
In addition, we provide \textit{enhanced experimental validation}, including extensive ablation studies dissecting each component (Section~\ref{sec:ablation},~\ref{sec:depth}), zero-shot generalization experiments on BDD100K~\cite{yu2020bdd100k} (Section~\ref{sec:BDD}) to demonstrate cross-domain transferability, and qualitative analysis (Section~\ref{sec:vis}) to verify each module's effectiveness.
Finally, the extended framework achieves \textit{improved performance}, with novel TETA improving from 34.3\% to 35.4\% (+1.1\%) on TAO validation and from 28.9\% to 30.5\% (+1.6\%) on TAO test, establishing new state-of-the-art performance.

\section{Related Work}
\label{sec:related}

\subsection{Multiple Object Tracking}
The dominant paradigm in MOT is the tracking-by-detection framework~\cite{andriluka2008people}, where detections are linked over time through association. Early studies emphasize appearance features~\cite{bergmann2019tracking, fischer2023qdtrack, leal2016learning, pang2021quasi, sadeghian2017tracking, wojke2017simple, cai2022memot} for re-identification, serving as the primary cue for object association.
Motion information also plays a critical role, with techniques utilizing Kalman Filtering~\cite{zhou2020tracking, saleh2021probabilistic, xiao2018simple, qin2023motiontrack, du2023strongsort} for trajectory prediction and 3D motion features~\cite{huang2023delving, luiten2020track, wang2023camo, krejvci2024pedestrian, ovsep2018track} for enhanced dynamics capture. However, motion-based methods like SORT~\cite{bewley2016simple} struggle with rapid motion and severe occlusions.
Hybrid approaches combine appearance and motion cues for robust tracking. DeepSORT~\cite{wojke2017simple} enhances appearance-based associations with motion priors, while dual-branch architectures~\cite{segu2024walker, wang2020towards, zhang2022bytetrack, zhang2021fairmot} and transformer-based methods~\cite{meinhardt2022trackformer, sun2020transtrack, zeng2022motr} integrate these cues through feature fusion, addressing challenges like occlusion. However, these methods focus primarily on simple categories (\textit{e.g.}, humans) and lack generalization to diverse object categories. Moreover, when extended to open-vocabulary scenarios, the varying reliability of different cues across novel categories presents new challenges for effective feature fusion.  

In this work, in addition to the commonly used appearance and motion, we also use category-aware semantic cues for OVMOT, which is specific to this problem and different from classical MOT. To effectively leverage different types of cues, we develop a \revised{confidence-guided feature fusion strategy} to dynamically balance and integrate the appearance, motion and semantic cues. Moreover, OVMOT introduces dense detection challenges with hierarchical spatial relationships and temporal detection instability, which we address through spatial feature aggregation and graph-based confidence propagation mechanisms.

\subsection{Open-World/Vocabulary MOT}
To expand the object categories in MOT, Dave et al.~\cite{dave2020tao} introduced the TAO benchmark to evaluate tracking under a long-tail class distribution. Methods like AOA~\cite{du20211st}, GTR~\cite{zhou2022global}, TET~\cite{li2022tracking}, and QDTrack~\cite{fischer2023qdtrack} perform generic object tracking on TAO. However, these methods are limited to predefined categories and cannot handle novel class objects absent from the training set.

To address this limitation, open-world MOT aims to detect and track objects not seen in the training set. Early works focus on class-agnostic detection and tracking~\cite{mitzel2012taking,ovsep2018track}, while the TAO-OW benchmark~\cite{liu2022opening} evaluates open-world tracking. However, TAO-OW relies on class-agnostic metrics, which fail to identify the specific classes of unknown objects.
Li et al.~\cite{li2023ovtrack} proposed a related problem, open-vocabulary multi-object tracking (OVMOT), where OVTrack extends the tracking framework to open-vocabulary settings. Similarly, MASA~\cite{Li_2024_CVPR} leverages unlabeled image pairs to learn universal appearance models. However, both OVTrack and MASA rely heavily on appearance-based strategies, limiting their generalization to novel categories. SLAck~\cite{li2024slack} advances the field by integrating semantic, motion, and appearance features, eliminating the need for heuristic post-processing. However, SLAck's feature fusion via simple summation operation fails to fully exploit the complementary potential of semantic and motion information.
More recently, OVSORT~\cite{li2025open}, OVTR~\cite{li2025ovtr}, and TRACT~\cite{li2025attention} extend OVMOT with improved architectures or training strategies. OVTR introduces a new transformer-based tracker, while OVSORT and TRACT explore domain generalization and trajectory-related cues. 

However, these methods are still trained on image datasets, lacking continuous temporal supervision for learning robust motion dynamics. Moreover, they do not employ adaptive mechanisms to balance multiple cues based on their varying reliability across different scenarios.
In contrast, this work proposes a more efficient feature fusion method with intra- and inter-frame confidence, integrating semantic and motion cues into appearance-based representations. Additionally, we introduce hierarchical spatial cross-attention to exploit parent-child relationships in dense detections and temporal graph propagation to address detection instability. The proposed C-TAO dataset further enhances tracking performance for both our and other methods in diverse open-vocabulary scenarios.

\section{C-TAO: Continuous TAO Dataset} 
\label{sec:dataset}

\subsection{Motivation: Training Challenges with TAO}

The TAO dataset emerges as a uniquely valuable video resource for training due to its extensive category coverage, vastly surpassing traditional MOT datasets limited to specific categories (person, vehicle, \etc.). However, its \textit{sparse annotation strategy}, which provides annotations every 30 frames, poses significant training challenges. The large temporal gaps hinder learning accurate motion patterns, as the model cannot observe smooth trajectories between annotated frames.
Additionally, the lack of intermediate frame annotations makes it difficult to capture fine-grained appearance changes and effectively handle occlusions. To compensate, researchers often supplement training with image datasets like LVIS~\cite{gupta2019lvis} and SA-1B~\cite{kirillov2023segment} constructing synthetic image pairs~\cite{li2023ovtrack, Li_2024_CVPR, li2025open,li2025attention} to simulate adjacent frames. However, these pairs fail to simulate continuous motion and realistic appearance changes, limiting model performance. Recent SLAck~\cite{li2024slack} generates pseudo labels via IoU matching to create continuous annotations in TAO, showing substantial improvements over image-pair-based methods. Yet, IoU matching lacks temporal consistency, especially in dynamic scenarios, leading to unreliable motion features.
To address these limitations, a continuously annotated version of the TAO training set is urgently needed. \revised{Such a dataset provides direct supervision for local temporal association and enables a clearer analysis of how dense trajectory annotations affect OVMOT training.}

\subsection{Construction of C-TAO Dataset}

\revised{To create a continuously annotated high-quality dataset while preserving the original data distribution, we retain all videos and annotated trajectories from the TAO training set, including the original diversity in categories, scenes, and motion patterns. C-TAO is a dense completion of the original TAO training trajectories rather than an exhaustive re-annotation of all visible objects. It keeps the original video IDs, frame IDs, track IDs, category labels, and sparse trajectory anchors unchanged, and only adds missing visible boxes along these existing trajectories. Fully occluded or out-of-frame targets are not hallucinated as visible boxes.}

\revised{The annotation process follows a completion-and-verification workflow. Annotators first complete the missing visible boxes using only the raw frames and the original sparse TAO boxes, track IDs, and category labels as anchors. A second pass then checks box tightness, temporal ID consistency, visibility decisions, and category consistency. Difficult cases, including heavy occlusion, re-entry, truncation, motion blur, and visually similar neighboring instances, are escalated to a senior annotator for adjudication. Before finalizing the annotations, we further run consistency checks on the COCO-video annotation files, including JSON schema validation, valid video/frame/track references, positive box sizes, image-boundary checks, duplicate annotation detection, preservation of original TAO track/category IDs, verification that the base-only training file contains no novel-class training labels, and flags for unexplained temporal gaps or abrupt visibility/box-size transitions. All flagged cases are manually revisited before finalizing the annotations. The base-only leakage check returns zero novel-class annotations in the strict base-only training file.}

\revised{No model-generated boxes, detector/tracker predictions, interpolation results, or outputs from COVTrack++ or any compared method are used for box initialization, identity association, trajectory completion, or verification. This prevents the completed annotations from being biased toward our method or any specific tracker.} 


\revised{The C-TAO annotation files, file-level documentation, and code for dataset preparation, training, inference, and evaluation are publicly available through the project repository and linked data package at \url{https://github.com/zekunqian/covtrack}.}

\subsection{Comparison and Analysis}

\subsubsection{Statistical Comparison}
\begin{figure}[!t]
    \centering
    \includegraphics[width=\columnwidth]{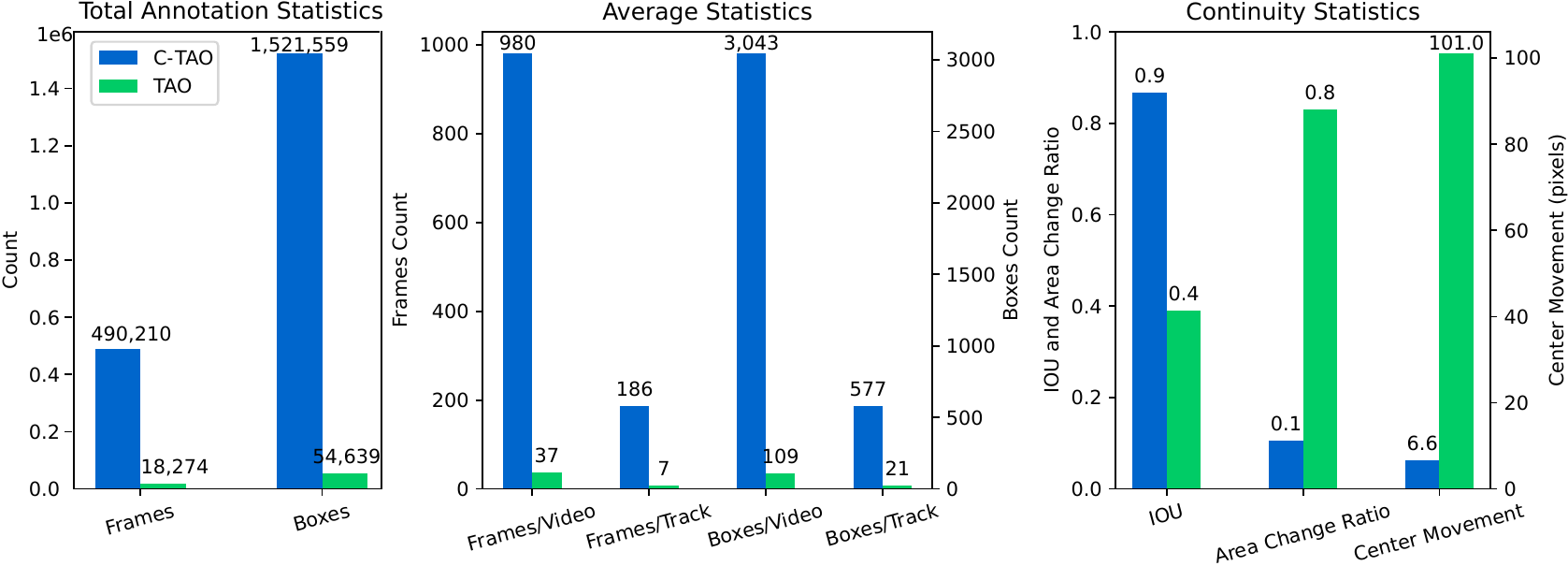}
    \caption{Visualization of annotation statistics comparison between our dataset (C-TAO) and TAO. Left: total number of annotated frames and bounding boxes. Middle: average statistics per video and per track. Right: continuity statistics between consecutive annotated frames.}
    \label{fig:compare_stats}
\end{figure}

We compare our C-TAO dataset with the original TAO dataset in terms of annotation density, coverage, and continuity, as shown in Fig.~\ref{fig:compare_stats}. Both datasets share the same videos and trajectories, and C-TAO significantly enhances annotation density, with total annotated frames increasing by 26$\times$ and total bounding boxes increasing by 27$\times$, as illustrated in the left subplot.
At the video level, C-TAO shows a dramatic improvement in annotated frame coverage and the number of annotations per video, ensuring a more comprehensive capture of object dynamics and scene evolution, as shown in the center subplot. For trajectory-level statistics, C-TAO exhibits denser temporal sampling, with remarkable increases in frames per track and annotations per track, providing more complete object motion information.
\revised{The right subplot further analyzes annotation continuity between consecutive annotated boxes along the same original TAO trajectory, including average IoU, area change ratio, and center movement. The results show much higher adjacent-box IoU and much smaller area/center changes, indicating that C-TAO significantly shortens the temporal gap between adjacent supervision signals. This provides dense short-baseline trajectory supervision, which is an important factor for improving tracking performance.}

\begin{figure}[!t]
    \centering
    \includegraphics[width=\columnwidth]{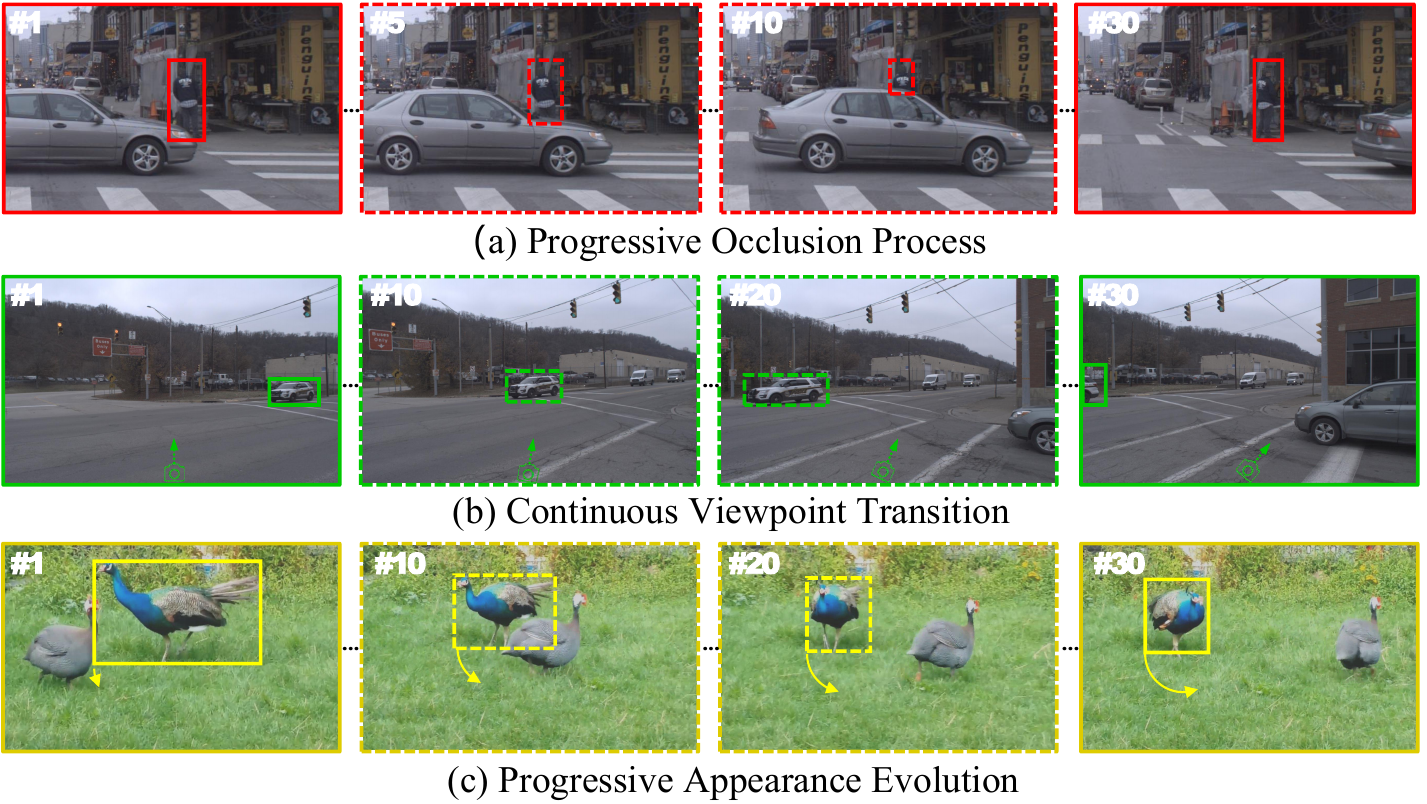}
    \caption{Annotation examples in challenging scenarios. Solid boxes represent original TAO annotations (30-frame intervals), while dashed boxes show our continuous annotations. (a) Progressive occlusion process of a pedestrian behind a car. (b) Continuous viewpoint transition of a vehicle under camera motion. (c) Progressive appearance evolution during a bird's pose transformation. C-TAO annotations capture crucial intermediate states that are missed in the original sparse annotations.}
    \label{fig:dense_annotation}
\end{figure}

\subsubsection{Qualitative Analysis}
To further demonstrate the advantages of C-TAO, we present three representative scenarios in Fig.~\ref{fig:dense_annotation}. These examples illustrate how our continuous annotations capture crucial intermediate states that are missing in the original TAO dataset's sparse annotations.
First, as shown in Fig.~\ref{fig:dense_annotation}(a), for the very common occlusion cases, our annotations capture the complete progression of the target moving through the occlusion, providing required training samples for handling complex occlusion scenes.
Second, as shown in Fig.~\ref{fig:dense_annotation}(b), under significant camera motion, the dense annotations reveal smooth viewpoint transitions that bridge the large perspective gaps present in sparse annotations.
Third, as shown in Fig.~\ref{fig:dense_annotation}(c), for objects undergoing appearance changes, our annotations record the continuous evolution of the target appearance, not only the object before and after the significant shape deformations.

Qualitative examples and statistical improvements demonstrate how our densely annotated data enhances the original TAO by providing more effective temporal information, which is crucial for MOT systems to learn robust tracking features, especially in challenging scenarios.

\subsection{\revised{Dense C-TAO Validation Split}}
\label{sec:ctao_val}

\revised{The conference version mainly used C-TAO as continuous training supervision. In this journal version, we further add C-TAO-val as a continuously annotated validation split by completing visible boxes along the original TAO validation trajectories. Its construction follows the same trajectory-completion and quality-control process as the C-TAO training annotations, but is applied to the original TAO validation trajectories. Compared with sparse TAO-val, C-TAO-val provides denser intermediate-frame annotations and enables more fine-grained evaluation of trajectory continuity, while the official sparse TAO validation/test protocol remains the primary comparison protocol with prior work.} 

\section{Proposed Method}
\label{sec:method}

\begin{figure*}[!t]
	\centering 
\includegraphics[width=1.0\textwidth]{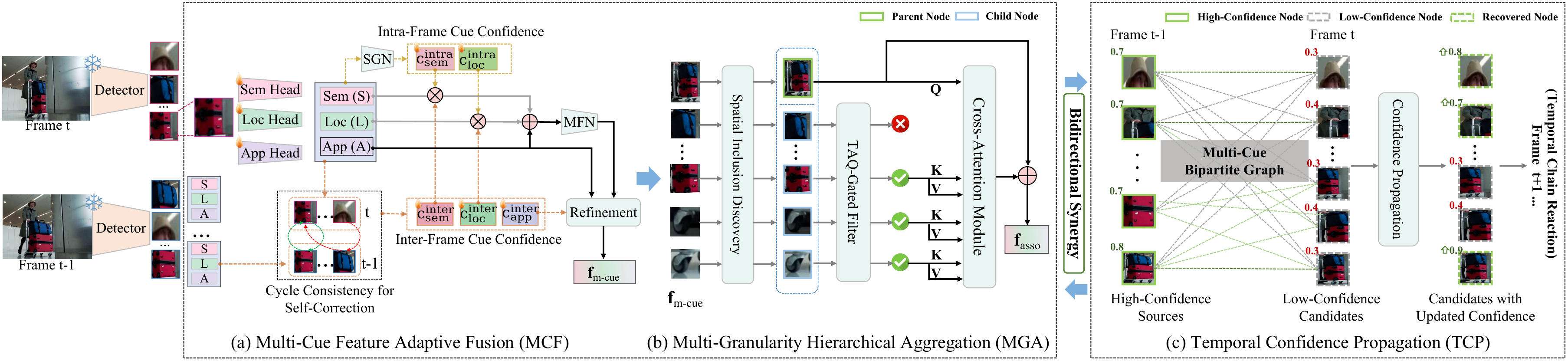}
	\caption{Overall framework of our method containing three complementary modules. (a) \textit{Multi-cue feature adaptive fusion (MCF)} dynamically weights appearance, location, and semantic features based on intra-frame and inter-frame confidence. (b) \textit{Multi-granularity hierarchical aggregation (MGA)} enhances parent objects (e.g., trolley) by aggregating spatial and semantic information from child parts (e.g., luggage, wheels) through cross-attention. (c) \textit{Temporal confidence propagation (TCP)} recovers low-confidence candidates in frame $t$ by leveraging high-confidence sources from frame $t-1$ via a bipartite graph matching. $\bigotimes$ and $\bigoplus$ represent scaling multiplication and concatenation operations.} 
	\label{fig:framework} 
\end{figure*}

\subsection{Overview}

We handle the OVMOT problem through two complementary aspects that mutually reinforce each other, as illustrated in Fig.~\ref{fig:framework}.
On the one hand, for the object association task, we aim to learn the effective association features through the multi-cue feature adaptive fusion and multi-granularity hierarchical aggregation.
On the other hand, based on the temporal association features, we further improve the detection quality using the cross-frame temporal guidance, which can effectively recover the missed detections.
This way, we have built a bidirectional reciprocal mechanism: the association features provide the prior for detection improvement (\textit{association to detection}), and the recovered detections provide more sources for the multi-cue multi-granularity association feature learning  (\textit{detection to association}).

\subsubsection{Association Feature Learning}
We construct robust association features from multiple cues and multiple granularities.
At the multi-cue stage, we apply a \textit{multi-cue adaptive fusion} (MCF) strategy.
Specifically, OVMOT provides three complementary cues: appearance, {motion (enabled by C-TAO's continuity)}, and semantics, yet their reliability varies across scenarios.
We therefore employ dual-perspective confidence estimation: intra-frame confidence learns mutual relationships among cues, while inter-frame confidence leverages temporal cycle consistency to evaluate feature stability, enabling adaptive fusion (Section~\ref{sec:multi_cue}).
At the multi-granularity stage, dense detections create parent-child structures where child parts can remain visible even when the parent is occluded.
We therefore introduce \textit{multi-granularity hierarchical aggregation (MGA)} to exploit these structures and enhance parent features with reliable child evidence (Section~\ref{sec:MGA}). 

\subsubsection{Detection Candidate Recovery}
We next address the detection flickering problem via \textit{temporal confidence propagation (TCP)}.
TCP builds a bipartite graph to construct the connections between low-confidence candidates in frame $t$ and high-confidence sources in frame $t-1$, and propagates confidence through reliable temporal correspondences to recover suppressed detections that would otherwise break trajectories (Section~\ref{sec:tcp}).

\subsubsection{Association-Detection Mutual Reinforcement}
Reliable association features from MCF and MGA can help recover the detections, and those recovered detections from TCP serve as high-quality tracking sources for subsequent frames.
This creates a bidirectional synergy: better association features improve detection recovery, which in turn yields stronger sources for future association, progressively improving both tracking continuity and detection quality.


\subsection{Multi-Cue Feature Adaptive Fusion (MCF)}\label{sec:multi_cue}

\subsubsection{Multi-Cue Feature Extraction}
\label{sec:feature_cues}
To ensure a fair comparison, we extract multi-cue features using the same pre-trained detector backbone as previous methods. Specifically, we adopt the open-vocabulary detector utilized in OVTrack~\cite{li2023ovtrack} and SLAck~\cite{li2024slack}, keeping its parameters frozen during training. Based on it, we construct the features for association from appearance, location (motion), and semantic cues.

\textit{\ding{202} Appearance head.}
To capture visual details for effective object association, we extract Region-of-interest (RoI) features from the detection proposals. These features are subsequently processed by a lightweight convolutional module followed by an MLP. The resulting transformation produces the $i$-th object's appearance embedding, $\mathbf{e}_{\text{app}}^{i} \in \mathbb{R}^{d}$.

\textit{\ding{203} Location (motion) head.}  
The location head encodes the spatial attributes of detected objects.
Given a bounding box, we normalize its coordinates with respect to the image dimensions.
The normalized values capture the object's relative spatial position in a scale-invariant manner, which are then input into a dedicated projection layer to yield the $i$-th object's location embedding, $\mathbf{e}_{\text{loc}}^{i} \in \mathbb{R}^{d}$.
Across frames, these location embeddings capture temporal consistency, which we use as a lightweight motion cue. For brevity, we also refer to this location-derived signal as the motion cue in the paper.

\textit{\ding{204} Semantic head.}
To produce versatile semantic representations without re-training, we initially consider leveraging CLIP~\cite{radford2021learning} but opt to distill its semantic head due to high inference costs. Following a distillation process similar to~\cite{gu2021open}, we fuse CLIP's text and image embeddings via element-wise summation, refine them with an MLP, and obtain the final $i$-th object's semantic embedding $\mathbf{e}_{\text{sem}}^{i} \in \mathbb{R}^{d}$. This dual-path design with both textual context and visual details for semantic features is verified in~\cite{gu2021open, li2023ovtrack}.

By explicitly modeling appearance, semantic cues, and location-derived motion cues, our framework constructs a rich feature representation for the open-vocabulary tracking (association) task.

\subsubsection{Confidence Estimation}
\label{sec:cue_confidence}
Appearance is the most commonly used feature in previous OVMOT and MOT tasks. While appearance features serve as a solid foundation, incorporating semantic and location cues can significantly enhance the discriminative capability. However, effectively fusing these features from different cues presents significant challenges. 

Given the three feature embeddings $\mathbf{e}_{\text{app}}^{i}$, $\mathbf{e}_{\text{loc}}^{i}$, and $\mathbf{e}_{\text{sem}}^{i}$ defined above, a straightforward fusion approach, as adopted in SLAck~\cite{li2024slack}, is direct summation: $
\mathbf{e}_{\mathrm{fused}}^{i}=\mathbf{e}_{\mathrm{app}}^{i}+\mathbf{e}_{\mathrm{loc}}^{i}+\mathbf{e}_{\mathrm{sem}}^{i}.
$
While simple, this equal-weighting strategy faces substantial challenges in OVMOT. Since object localization and classification are inherently difficult tasks, the reliability of location and semantic cues varies significantly across different scenarios. This reliability variation becomes particularly problematic for novel categories, where classification accuracy often remains in single digits. Consequently, such naive feature fusion can introduce considerable noise and destabilize tracking performance.

To address these fusion reliability issues, we propose a dual-perspective confidence estimation strategy that dynamically determines the contribution weights of different cues. Our approach evaluates feature reliability from two complementary perspectives:
\textit{intra-frame confidence} assesses the reliability of semantic and location features within a single frame by learning their mutual relationships with appearance features, while \textit{inter-frame confidence} leverages temporal consistency between adjacent frames to evaluate the stability of each feature type across time.
This dual-perspective design enables adaptive feature weighting that accounts for both spatial and temporal reliability variations.

\textit{\ding{202} Intra-frame cue confidence.}
The $i$-th object's intra-frame cue confidence, denoted as $c^{\text{intra}}_{\text{sem},i}$ and $c^{\text{intra}}_{\text{loc},i}$, are computed using the Self-attentive gated network (SGN). This network evaluates the relative reliability of the semantic and location cues within a single frame, ensuring that their impacts on the final fused feature are adaptively weighted.

As shown in Fig.~\ref{fig:framework}(a), on the current frame $t$, the network takes three cues as input, which interact at the feature level through concatenation operations.
Then the network learns the \textit{mutual confidence relationships} between these cues to construct effective intra-frame confidence scores of the $i$-th object as 
\begin{equation}
	\label{eq:SGN}
	[c^{\text{intra}}_{\text{loc},i}, c^{\text{intra}}_{\text{sem},i}] = {\rm{sigmoid}} \left({\text{SGN}}( \text{concat}(\textbf{e}_{\text{app}}^{i}, \textbf{e}_{\text{loc}}^{i}, \textbf{e}_{\text{sem}}^{i}))\right),
\end{equation}
where SGN is a Self-attentive gated network, and the sigmoid activation ensures that the gate values $c^{\text{intra}}_{\text{loc},i}$ and $c^{\text{intra}}_{\text{sem},i}$ are within the range $[0, 1]$. 
\revised{Eq.~\myeqref{eq:SGN} estimates the intra-frame confidence scores from the concatenated cues. Let $\mathbf{x}_i=[\mathbf{e}^{i}_{\mathrm{app}};\mathbf{e}^{i}_{\mathrm{loc}};\mathbf{e}^{i}_{\mathrm{sem}}]$ denote the concatenated cue feature. SGN parameterizes this multi-cue mapping as $\mathbf{h}_i=\phi(\mathbf{W}_1\mathbf{x}_i+\mathbf{b}_1)$ and $[c^{\mathrm{intra}}_{\mathrm{loc},i},c^{\mathrm{intra}}_{\mathrm{sem},i}]=\sigma(\mathbf{W}_2\mathbf{h}_i+\mathbf{b}_2)$.
Consequently, both gates are predicted from the same multi-cue representation and are conditioned jointly on appearance, location, and semantic cues, rather than being independent hand-designed scalar weights.} 

These intra-frame confidence scores \textit{dynamically adjust the relative influence of the location and semantic cues} in the fused feature.
Higher confidence values indicate that the corresponding cue is more reliable and should contribute more to the final fused feature.
This mechanism ensures that the fusion process is robust to noise and uncertainty, from a single-frame level.  

\textit{\ding{203} Inter-frame cue confidence.}
To evaluate the temporal reliability of different cues across continuous frames, as shown in Fig.~\ref{fig:framework}(a), we design an inter-frame cycle consistency estimation module.
Given appearance, location, and semantic features, we take the appearance cue as an example. Let $\mathbf{E}^t_{\text{app}} \in \mathbb{R}^{n \times d}$ and $\mathbf{E}^{t-1}_{\text{app}} \in \mathbb{R}^{m \times d}$ represent the feature matrices at frames $t$ and $t-1$, where $n$ and $m$ denote the number of objects in each frame, and $d$ is the feature dimension. We compute the pairwise similarity matrix between objects in consecutive frames as
$
\mathbf{S}_{\text{app}} = \mathbf{E}^t_{\text{app}} \cdot \left(\mathbf{E}^{t-1}_{\text{app}}\right)^\top \in \mathbb{R}^{n \times m}.
$
To enhance the discriminative power of similarity scores, we apply adaptive temperature scaling with parameter $\alpha$.
\revised{The cycle operation is formulated with two separately row-normalized transition matrices:
\begin{equation}
	\mathbf{P}^{t\rightarrow t-1}_{\text{app}}
	=
	\operatorname{softmax}_{\mathrm{row}}
	\left(\alpha \mathbf{E}^{t}_{\text{app}}(\mathbf{E}^{t-1}_{\text{app}})^\top\right),
\end{equation}
\begin{equation}
	\mathbf{P}^{t-1\rightarrow t}_{\text{app}}
	=
	\operatorname{softmax}_{\mathrm{row}}
	\left(\alpha \mathbf{E}^{t-1}_{\text{app}}(\mathbf{E}^{t}_{\text{app}})^\top\right).
\end{equation}
The cycle-confidence matrix is
\begin{equation}
	\label{eq:cycle}
	\mathbf{C}_{\text{app}}^{\text{cycle}}
	=
	\mathbf{P}^{t\rightarrow t-1}_{\text{app}}
	\mathbf{P}^{t-1\rightarrow t}_{\text{app}}.
\end{equation}
Following cycle-consistency based correspondence learning~\cite{wang2019learning}, for object $i$ in frame $t$, its inter-frame appearance confidence is the return probability of this two-step random walk:
\begin{equation}
	c^{\text{inter}}_{\text{app},i}
	=
	\left[\operatorname{diag}\left(\mathbf{C}_{\text{app}}^{\text{cycle}}\right)\right]_i
	=
	\sum_k
	P^{t\rightarrow t-1}_{\text{app},ik}
	P^{t-1\rightarrow t}_{\text{app},ki}.
\end{equation}
This differs from the self-product diagonal
$[\hat{\mathbf{S}}\hat{\mathbf{S}}^\top]_{ii}=\sum_k\hat{S}_{ik}^{2}$, which only measures the second-order concentration of a one-way row-softmax distribution. Because the reverse transition is separately row-normalized over current-frame objects rather than obtained by transposing the forward row-softmax, the score couples two opposite-direction conditional transitions and introduces reverse-direction competition that suppresses one-sided sharp matches and multi-to-one drifts. Thus, higher cycle confidence indicates stronger reciprocal temporal stability, while lower confidence indicates ambiguous or unstable cue matching.}
Similarly, this pipeline can be applied to location and semantic features to obtain $c^{\text{inter}}_{\text{loc},i}$ and $c^{\text{inter}}_{\text{sem},i}$ respectively.

\subsubsection{Multi-Cue Feature Fusion}
\label{sec:cue_fusion}
With the above confidences, for the $i$-th object, we construct a multi-cue feature representation by combining the original appearance feature with confidence-weighted location and semantic features, which are modulated by their corresponding intra-frame and inter-frame confidence scores as
\begin{equation}
	\tilde{\textbf{e}}_{\text{loc}}^{i} = c^{\text{intra}}_{\text{loc},i} \cdot c^{\text{inter}}_{\text{loc},i} \cdot \textbf{e}_{\text{loc}}^{i}, \quad
	\tilde{\textbf{e}}_{\text{sem}}^{i} = c^{\text{intra}}_{\text{sem},i} \cdot c^{\text{inter}}_{\text{sem},i} \cdot \textbf{e}_{\text{sem}}^{i}.
\end{equation}
These features are concatenated with appearance to form a multi-cue representation, which is input to a Multi-cue fusion network (MFN) to integrate information from all cues and output a feature with the original embedding dimension
\begin{equation}
	\label{eq:m-cue}
	{{\textbf{f}}}^{i}_{\text{fus}} = \text{MFN}\big(\text{concat}\big({\textbf{e}}_{\text{app}}^{i},\, \tilde{{\textbf{e}}}_{\text{loc}}^{i},\, \tilde{{\textbf{e}}}_{\text{sem}}^{i}\big)\big) \in \mathbb{R}^{d},
\end{equation}
where the fused feature ${{\textbf{f}}}^{i}_{\text{fus}}$ effectively encodes the appearance, location, and semantic information with their intra-/inter-frame relationships.

Finally, we apply a \textit{feature refinement strategy with re-connection.} In this refinement step, we combine the multi-cue fused feature ${{\textbf{f}}}^{i}_{\text{fus}}$ with the main appearance feature ${\textbf{e}}_{\text{app}}^{i}$.
Specifically, using the inter-frame confidence score of appearance features $c^{\text{inter}}_{\text{app},i}$ as guidance, we adaptively combine ${{\textbf{f}}}^{i}_{\text{fus}}$ with ${\textbf{e}}_{\text{app}}^{i}$ as the final multi-cue feature ${{\textbf{f}}}_{\text{m-cue}}^{i}$ for association in OVMOT:
\begin{equation}
	\label{eq:refine}
	{{\textbf{f}}}_{\text{m-cue}}^{i} = c^{\text{inter}}_{\text{app},i} \cdot {\textbf{e}}_{\text{app}}^{i} + (1 - c^{\text{inter}}_{\text{app},i}) \cdot {{\textbf{f}}}^{i}_{\text{fus}} \in \mathbb{R}^{d}.
\end{equation}

This refinement mechanism adaptively balances between the original appearance feature and the aggregated feature based on the \textit{appearance self-correction}. When appearance features demonstrate strong temporal consistency (high $c^{\text{inter}}_{\text{app},i}$), the model places greater emphasis on ${\textbf{e}}_{\text{app}}^{i}$. Conversely, when appearance features show weak temporal consistency (low $c^{\text{inter}}_{\text{app},i}$), the model relies more on the information-rich aggregated feature $ {{\textbf{f}}}^{i}_{\text{fus}}$ which incorporates multiple complementary cues. This adaptive mechanism ensures robust feature fusion by dynamically adjusting the impact of each cue.

\subsubsection{Training Method and Analysis}
\label{sec:loss}
To train the feature fusion framework, we compute the similarity between objects in consecutive frames using the final multi-cue fused features, \ie, ${{\textbf{F}}}_{\text{m-cue}}^t \in \mathbb{R}^{n \times d}$ and ${{\textbf{F}}}_{\text{m-cue}}^{t-1} \in \mathbb{R}^{m \times d}$, as
$
\mathbf{S}  = {{\textbf{F}}}_{\text{m-cue}}^t \cdot ({{\textbf{F}}}_{\text{m-cue}}^{t-1})^\top \in \mathbb{R}^{n \times m}.
$
The multi-cue loss is formulated as
\begin{equation}
	\label{eq:asso}
	\mathcal{L}_{\text{m-cue}} = -\textstyle\sum_{i,j} \mathbf{Y}_{ij}\log(\hat{\mathbf{S}}_{ij}),  i \in \{1,\ldots,n\}, j \in \{1,\ldots,m\},
\end{equation}
where $\mathbf{Y}$ is the ground-truth association label, and $\hat{\mathbf{S}}$ represents the row-wise normalized $\mathbf{S}$ via softmax.

{
	The multi-cue loss enables end-to-end training of all confidence scores through backpropagation.
	Specifically, five learnable confidence scores are trained in different stages of feature fusion.
	During training, these scores $c$ are optimized through the continuous gradient flow as
	$
	\frac{\partial \mathcal{L}_{\text{m-cue}}}{\partial c} = \frac{\partial \mathcal{L}_{\text{m-cue}}}{\partial \mathbf{S}} \cdot \frac{\partial \mathbf{S}}{\partial {{\textbf{f}}}_{\text{m-cue}}} \cdot \frac{\partial {{\textbf{f}}}_{\text{m-cue}}}{\partial {{\textbf{f}}}_{\text{fus}}} \cdot \frac{\partial {{\textbf{f}}}_{\text{fus}}}{\partial \tilde{\textbf{e}}} \cdot \frac{\partial \tilde{\textbf{e}}}{\partial c}.
	$
	\revised{This gradient path passes through MFN, the gate-application step that forms \(\tilde{\textbf{e}}\), and the SGN confidence-estimation modules, so the reliability gates and fused association features are jointly optimized under \(\mathcal{L}_{\text{m-cue}}\) rather than tuned as post-hoc scalar weights.}
	This fully differentiable design ensures that all confidence scores can be effectively trained.
	Through back-propagation, all the confidence weights are \textit{learned to be optimally balanced among different cues, without any explicit supervision}.
	The end-to-end training allows these scores to automatically adapt to the mutual relation of different cues and the temporal self-correction along the video.}

\subsection{Multi-Granularity Hierarchical Aggregation (MGA)}
\label{sec:MGA}
The multi-cue adaptive fusion framework produces a robust feature representation $\mathbf{f}_{\text{m-cue}}^i \in \mathbb{R}^d$ for each detected object $i$.
However, these features are computed independently and overlook spatial relationships among dense detections.
As illustrated in Fig.~\ref{fig:framework}(b), open-vocabulary detection produces nested bounding boxes, where a parent object (\eg, trolley) often contains multiple child regions (\eg, luggage, wheels).
These {hierarchical structures encode multi-granularity information that can be used for representation}: children provide fine-grained local details, while the parent encodes holistic semantics.
Under partial occlusion, visible child components remain stable and can sustain the parent representation.
To exploit this, we propose MGA, which leverages spatial inclusion relationships to enhance parent features by selectively aggregating reliable child evidence.
\revised{This design follows a reliability-aware multi-observation association principle: spatially contained child regions provide auxiliary evidence for the same parent identity, while MGA still performs object-level association rather than independent part tracking or category-specific aggregation.}

\subsubsection{Spatial Inclusion-Relation Mask}
Given $N$ detected objects in frame $t$ with bounding boxes $\{\mathbf{b}_i\}_{i=1}^N$ and their corresponding features $\{\mathbf{f}_{\text{m-cue}}^i\}_{i=1}^N$ from the multi-cue fusion module, we first identify hierarchical spatial relationships.
Two objects $i$ and $j$ exhibit a parent-child relationship when box $\mathbf{b}_j$ is spatially contained within box $\mathbf{b}_i$.
We formalize this using the Intersection over child (IoC) metric:
\begin{equation}
	\text{IoC}(i, j) = \frac{\text{Area}(\mathbf{b}_i \cap \mathbf{b}_j)}{\text{Area}(\mathbf{b}_j)}.
\end{equation}
If $\text{IoC}(i, j) > \tau_{\text{ioc}}$, box $j$ is considered a child of box $i$.
Based on this, we construct a Spatial Inclusion Mask $\mathbf{M} \in \{0,1\}^{N \times N}$ to encode parent-child relationships:
\begin{equation}
	\mathbf{M}_{ij} =
	\begin{cases}
		1, & \text{if IoC}(i, j) > \tau_{\text{ioc}} \text{ and } i \neq j \\
		0, & \text{otherwise}
	\end{cases}.
\end{equation}
Here, $\mathbf{M}_{ij} = 1$ indicates that object $j$ is a child of object $i$.
For each parent object $i$, the set of its children is $\mathcal{C}_i = \{j \mid \mathbf{M}_{ij} = 1\}$.

\subsubsection{TAQ-Gated Cross-Attention}
Not all detected objects are reliable, as false positives or low-quality detections may contaminate parent features.
\revised{Inspired by the inter-frame self-correction strategy in Eq.~(\ref{eq:cycle}), we introduce Temporal association quality (TAQ) scores to measure detection reliability through the same forward-backward cycle consistency.}
\revised{For each object \(i\), we compute its TAQ score \(q_i \in [0,1]\) by evaluating feature consistency between consecutive frames \(t-1\) and \(t\)}
\begin{equation}
	\revised{q_i =
	\left[
	\operatorname{diag}
	\left(
	\mathbf{P}^{t\rightarrow t-1}
	\mathbf{P}^{t-1\rightarrow t}
	\right)
	\right]_i,}
\end{equation}
\revised{where
\(\mathbf{P}^{t\rightarrow t-1}=
\operatorname{softmax}_{\mathrm{row}}
\left(\alpha\mathbf{F}_{\text{m-cue}}^t(\mathbf{F}_{\text{m-cue}}^{t-1})^\top\right)\)
and
\(\mathbf{P}^{t-1\rightarrow t}=
\operatorname{softmax}_{\mathrm{row}}
\left(\alpha\mathbf{F}_{\text{m-cue}}^{t-1}(\mathbf{F}_{\text{m-cue}}^t)^\top\right)\)
are separately row-normalized transition matrices, following Eq.~(\ref{eq:cycle}).}
\revised{$\alpha$ is the temperature scaling parameter (same as in Section~\ref{sec:cue_confidence}).}
Higher TAQ scores indicate temporally stable, reliable detections, while low scores suggest spurious or unstable detections.

With the spatial inclusion mask $\mathbf{M}$ and TAQ scores, we apply cross-attention to aggregate child features for parent enhancement.
This process is performed in parallel for all parent objects.
For each parent object $i$, the spatial inclusion mask $\mathbf{M}$ determines its child set $\mathcal{C}_i = \{j \mid \mathbf{M}_{ij} = 1\}$.
Let $\mathbf{F}_{\mathcal{C}_i} = [\mathbf{f}_{\text{m-cue}}^j]_{j \in \mathcal{C}_i} \in \mathbb{R}^{|\mathcal{C}_i| \times d}$ denote the feature matrix of children, and $\mathbf{q}_{\mathcal{C}_i} = [q_j]_{j \in \mathcal{C}_i} \in \mathbb{R}^{|\mathcal{C}_i|}$ denote their TAQ scores.

We first apply TAQ-based gating to filter the input child features.
We construct a quality gate $\mathbf{g}_{\mathcal{C}_i} = \mathbf{I}(\mathbf{q}_{\mathcal{C}_i} > \tau_q) \in \{0,1\}^{|\mathcal{C}_i|}$, where $\mathbf{I}(\cdot)$ is the indicator function and $\tau_q$ is the quality threshold.
The gated child features are obtained by
\begin{equation}
	\tilde{\mathbf{F}}_{\mathcal{C}_i} = \mathbf{F}_{\mathcal{C}_i} \odot (\mathbf{g}_{\mathcal{C}_i} (\mathbf{1}_{d})^\top),
\end{equation}
where $\odot$ denotes element-wise multiplication and $\mathbf{1}_{d} \in \mathbb{R}^{d}$ is an all-ones vector.
This gating operation suppresses features of unreliable children (with $q_j \leq \tau_q$) by setting them to 0.

Next, we apply standard cross-attention: the parent feature serves as the query, while the gated child features serve as keys and values.
We compute the query, key, and value representations
\begin{equation}
	\mathbf{Q}_i = \mathbf{W}_q \mathbf{f}_{\text{m-cue}}^i, \quad \mathbf{K}_{\mathcal{C}_i} = \mathbf{W}_k (\tilde{\mathbf{F}}_{\mathcal{C}_i})^\top, \quad \mathbf{V}_{\mathcal{C}_i} = \mathbf{W}_v (\tilde{\mathbf{F}}_{\mathcal{C}_i})^\top,
\end{equation}
where $\mathbf{W}_q, \mathbf{W}_k, \mathbf{W}_v \in \mathbb{R}^{d \times d}$ are learnable projection matrices.
The attention scores are computed as
\begin{equation}
	\mathbf{s}_i = \frac{\mathbf{Q}_i^\top \mathbf{K}_{\mathcal{C}_i}}{\sqrt{d}} \in \mathbb{R}^{|\mathcal{C}_i|}.
\end{equation}
The child selection already applies the spatial inclusion mask, so the attention weights are computed via softmax
$
\mathbf{a}_i = \operatorname{softmax}(\mathbf{s}_i) \in \mathbb{R}^{|\mathcal{C}_i|}.
$
Finally, we aggregate the gated child features using these attention weights
\begin{equation}
	\mathbf{f}_\text{m-gra}^i = \mathbf{V}_{\mathcal{C}_i} \mathbf{a}_i^\top \in \mathbb{R}^{d}.
\end{equation}

The parent feature is enhanced by adding the aggregated child information through a residual connection.
To preserve the discriminability of leaf nodes (objects with no children), we only apply enhancement to parent objects
\begin{equation}
	\mathbf{f}_{\text{asso}}^{i} =
	\begin{cases}
		\mathbf{f}_{\text{m-cue}}^i + \lambda \cdot \mathbf{f}_\text{m-gra}^i, & \text{if } |\mathcal{C}_i| > 0 \\
		\mathbf{f}_{\text{m-cue}}^i, & \text{otherwise}
	\end{cases},
\end{equation}
where $\lambda$ is the enhancement ratio.
In summary, the spatial inclusion mask $\mathbf{M}$ acts as a structural mask in the cross-attention mechanism to select children based on spatial containment, while TAQ scores gate the input child features to filter out unreliable detections, ensuring that only stable, high-quality child features contribute to parent enhancement.

{
	MGA operates as a feature enhancement module that refines the multi-cue feature $\mathbf{f}_{\text{m-cue}}^i$ obtained in Section~\ref{sec:multi_cue} for the association.
	The enhanced features $\mathbf{f}_{\text{asso}}^{i}$ are used in the association loss in Eq.~(\ref{eq:asso}) for end-to-end training.
	The learnable projection matrices $\mathbf{W}_q, \mathbf{W}_k, \mathbf{W}_v$ are jointly optimized through the association loss, enabling the cross-attention mechanism to adaptively learn hierarchical parent-child relationships for tracking. 
}

\subsection{Temporal Confidence Propagation for Detection (TCP)}
\label{sec:tcp}
In the above sections, we have developed the multi-cue feature adaptive fusion, and multi-granularity feature hierarchical aggregation strategies for {reliable association representation learning}.
Besides association, OVMOT also \textit{faces challenges from the detection (localization) instability}.
Note that, as a video task, especially with the enhancement of the proposed continuous dataset, the detection can be improved through the temporal information from the continuous video.
This way, in this section, we further consider {improving the object localization results with the help of temporal association}. 
For this purpose, we propose a detection confidence temporal adjustment strategy to address the detection flickering by recovering the suppressed detections.

\begin{figure}[t]
	\centering
	\includegraphics[width=0.48\textwidth]{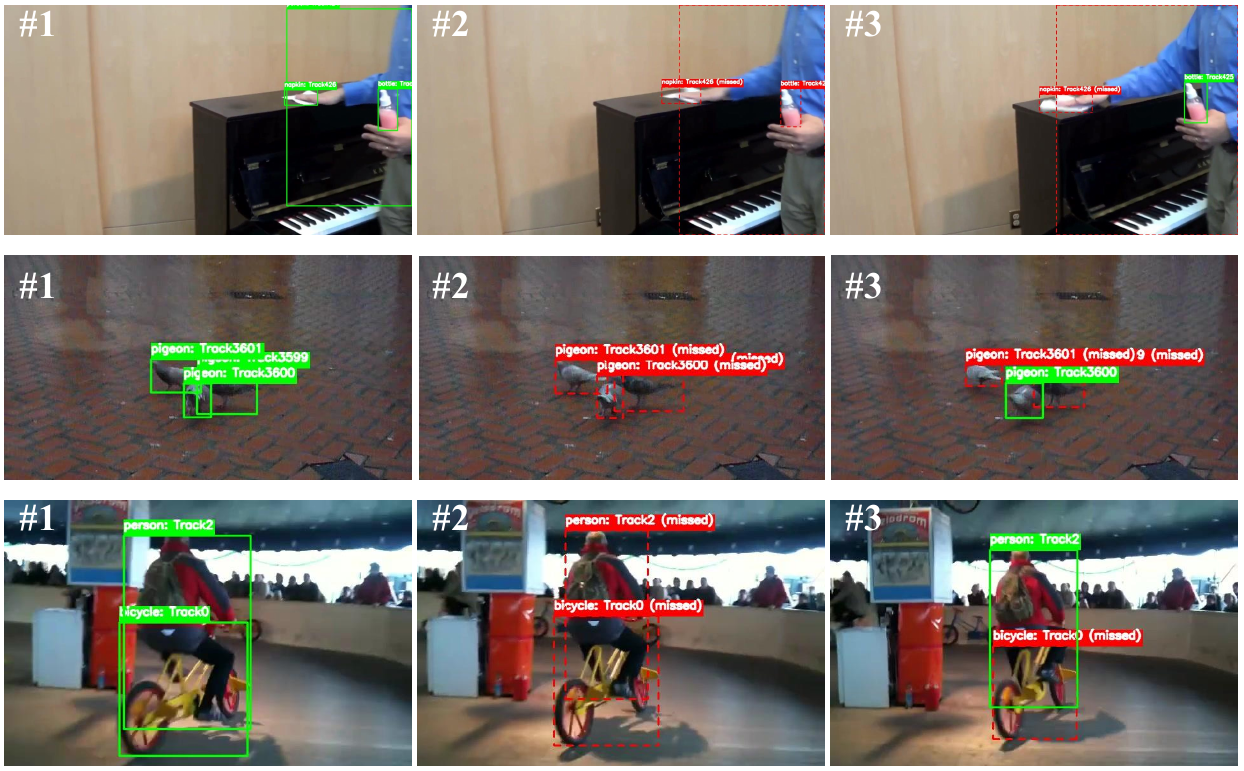}
	\caption{Illustration of detection flickering causing trajectory discontinuity. Green boxes indicate successful detections with high confidence, while red dashed boxes show missed detections due to confidence fluctuations. Despite minimal visual changes between consecutive frames, confidence score variations lead to detection flickering and fragmented trajectories.}
	\label{fig:flickering}
\end{figure}

Specifically, the detection process in OVMOT commonly follows a two-stage pipeline: the first stage employs salient object detection to propose as many potential targets as possible, while the second stage applies CLIP-distilled classification to identify targets with \textit{high classification confidence from these proposals to generate the detection boxes}.
However, achieving open-vocabulary recognition requires classification over an extremely large number of categories (e.g., LVIS class list contains 1,203 categories).
This classification complexity introduces a critical challenge: even minor appearance changes (such as slight motion blur or subtle viewpoint shifts) can cause significant fluctuations in detection confidence scores.
In other words, an object that produces a high-confidence detection in one frame may only generate a low-confidence detection in the next frame due to such subtle variations.
These low-confidence detections are then filtered out (as only detections above $\tau_{\text{high}}$ are retained), leading to \textit{detection flickering} and fragmented trajectories where objects flicker in and out of existence, as shown in Fig.~\ref{fig:flickering}.
Although these objects appear visually similar across adjacent frames, the detector fails to consistently produce high-confidence detections, resulting in trajectory discontinuity. This highlights the necessity of leveraging inter-frame associations to improve tracking robustness.

To address this, we propose TCP, a framework that recovers valid but low-confidence detections by exploiting temporal consistency with high-confidence tracked objects, as illustrated in Fig.~\ref{fig:framework}(c).
Specifically, in frame $t-1$ we apply standard NMS (IoU threshold 0.5) to obtain high-precision detections with confidence scores above $\tau_{\text{high}}$, which serve as reliable tracking sources.
Subsequently, we construct a bipartite graph to build the connection between low-confidence detection candidates (with confidence scores in $[\tau_{\text{low}}, \tau_{\text{high}}]$) in frame $t$ and these high-confidence tracked source detections from frame $t-1$. This way, we \textit{use a confidence propagation strategy to adjust the confidence} of detection candidates at frame $t$, which \textit{recovers} those valid but temporarily suppressed detection candidates. 

\subsubsection{Bipartite Graph for Cross-Frame Connection}
To achieve effective recovery of low-confidence detections, we need to establish robust cross-frame associations between candidates and tracked sources.
However, the dense detection nature of OVMOT introduces a significant challenge: for each low-confidence candidate, there may exist multiple potential matches from the previous frame's tracked objects, creating ambiguity in determining the correct association. 
To address this, we adopt a two-step approach.
First, we construct a multi-cue bipartite graph that models the temporal relationships between low-confidence candidates and high-confidence sources by integrating aggregated association features and location-based motion overlap.
Second, we develop a confidence propagation mechanism that leverages these multi-cue edges to transfer reliability from high-confidence tracked objects to their associated candidates. 

Formally, let $\mathcal{D}_{\text{source}}^{t-1} = \{d_i\}_{i=1}^{N_s}$ denote the $N_s$ high-confidence detections from frame $t-1$, with features $\{\mathbf{f}_{\text{asso}}^i\}_{i=1}^{N_s}$ (enhanced by MGA) and detection scores $\{s_i\}_{i=1}^{N_s}$.
Let $\mathcal{D}_{\text{cand}}^t = \{d_j\}_{j=1}^{N_c}$ denote the $N_c$ low-confidence candidates in the current frame $t$, with features $\{\mathbf{f}_{\text{asso}}^{j}\}_{j=1}^{N_c}$ and scores $\{s_j\}_{j=1}^{N_c}$ where $s_j \in [\tau_{\text{low}}, \tau_{\text{high}}]$.
We construct a bipartite graph $\mathcal{G} = (\mathcal{D}_{\text{source}}^{t-1}, \mathcal{D}_{\text{cand}}^t, \mathcal{E})$, 
where the two node sets represent reliable tracking sources from the previous frame, 
and the low-confidence candidates in the current frame, respectively.
Edges $\mathcal{E}$ encode multi-cue temporal relationships between them.

For each candidate $d_j$ and source $d_i$, we compute two types of edge weights representing different matching factors.
First, we measure the holistic feature similarity based on the aggregated association features $\mathbf{f}_{\text{asso}}^{i}$ (which encode multi-cue and multi-granularity information as discussed before)
\begin{equation}
	\revised{w_{ij}^{\mathrm{cos}} = \frac{\mathbf{f}_{\text{asso}}^{i} \cdot \mathbf{f}_{\text{asso}}^{j}}{\|\mathbf{f}_{\text{asso}}^{i}\| \|\mathbf{f}_{\text{asso}}^{j}\|},}
\end{equation}
capturing the overall similarity between the candidate and tracked source through cosine similarity with normalization. 
\revised{Since cosine similarity lies in $[-1,1]$, we calibrate it to $[0,1]$ before combining it with IoU:
\begin{equation}
	\bar{w}_{ij}^{\mathrm{asso}}=\frac{w_{ij}^{\mathrm{cos}}+1}{2}.
\end{equation}}

Second, besides the learnable features, we also measure motion consistency by directly computing the spatial overlap between the source's bounding box $\mathbf{b}_i^{t-1}$ in frame $t-1$ and the candidate's bounding box $\mathbf{b}_j^t$ in frame $t$,
\begin{equation}
	w_{ij}^{\mathrm{iou}} = \mathrm{IoU}(\mathbf{b}_i^{t-1}, \mathbf{b}_j^t).
\end{equation}
The final edge weight combines both cues through a \revised{normalized} weighted sum as
\begin{equation}
	\label{eq:w}
	\revised{w_{ij} =
	\frac{\bar{w}_{ij}^{\mathrm{asso}} + \beta w_{ij}^{\mathrm{iou}}}{1+\beta},}
\end{equation}
where 
$\beta$ is the weighting parameter. 
\revised{This calibration keeps $w_{ij}\in[0,1]$, making the evidence thresholds used for confidence propagation well-defined.}

\subsubsection{Class-Conditional Confidence Propagation}
Given the bipartite graph with edge weights $\{w_{ij}\}$, we now propagate confidence from high-quality tracked sources to low-confidence candidates.
The core challenge is that naively propagating confidence can introduce errors, especially for novel categories where the detector produces unreliable classification scores.
Our key insight is to make the propagation \textit{class-conditional}: we only propagate confidence between detections of the same predicted category, preventing semantic confusion across different object classes.
Although the absolute predicted category may be incorrect for novel objects (as reflected by low ClsA scores), the key observation is that the \textit{same object in adjacent frames tends to be assigned the same predicted category consistently}, even if that category is wrong.
This temporal consistency in predicted labels enables effective confidence propagation from a consistency perspective rather than relying on classification accuracy.
As will be demonstrated in our ablation study (Section~\ref{sec:ablation}), the class-conditional constraint effectively filters out spurious cross-category matches and significantly improves detection recovery quality, particularly for novel categories.

Let ${c}_i$ denote the predicted category of source $d_i$ and ${c}_j$ denote the predicted category of candidate $d_j$.
For each candidate $d_j$ with predicted class ${c}_j$, we identify the set of valid sources that share the same predicted category
\begin{equation}
	\mathcal{N}_j = \{i \mid {c}_i = {c}_j\},
\end{equation}
where $\mathcal{N}_j$ is a set containing the \textit{indices of sources} $i$ that match candidate $d_j$ in terms of predicted category.
The class-matching constraint ${c}_i = {c}_j$ ensures that confidence is only propagated within the same category, preventing spurious cross-category matches.

The adjusted confidence $\tilde{s}_j$ for candidate $d_j$ (with predicted class ${c}_j$) is computed by blending its original confidence $s_j$ with the weighted average confidence from matched sources as
\begin{equation}
	\label{eq:sj}
	\tilde{s}_j = (1 - \eta_j) \cdot s_j + \eta_j \cdot \frac{\sum_{i \in \mathcal{N}_j} w_{ij} \cdot s_i}{\sum_{i \in \mathcal{N}_j} w_{ij}},
\end{equation}
where $s_j$ denotes the original classification confidence of candidate $d_j$ for its predicted class ${c}_j$, $s_i$ is the confidence of source $d_i$ for the same class ${c}_j$, and $w_{ij}$ is the combined edge weight in Eq.~(\ref{eq:w}).
The propagation strength $\eta_j \in [0, 1]$ controls the balance between the original confidence and propagated confidence, adaptively determined based on the evidence strength.
We assign value to $\eta_j$ through two factors: the number in $\mathcal{N}_j$, \ie, $|\mathcal{N}_j|$, and the average value of all $w_{ij}$ with $i \in \mathcal{N}_j$, \ie, $\bar{w}_j$.
Details for the setting of $\eta_j$ are provided in the implementation details in Section~\ref{sec:imp_details}. 
If $\mathcal{N}_j$ is empty, we set $\eta_j = 0$ and keep the original score.
We assign higher propagation strength to candidates with strong evidence (high average edge weights and multiple supporting sources), while conservatively handling candidates with weak or ambiguous matches.

After obtaining the boosted class-specific confidence $\tilde{s}_j$, we update the candidate's overall detection score as $\tilde{s}_j = \max(\tilde{s}_j, s_j)$ since we aim to recover the missed low-confidence detections.
A candidate $d_j$ is recovered and added to the final detection set if its boosted confidence exceeds the detection threshold, \ie, 
$d_j \text{ is recovered if } \tilde{s}_j > \tau_{\text{high}}$.
Based on the above strategy, the recovered candidate inherits its boosted confidence score $\tilde{s}_j$ as its new detection confidence, enabling it to serve as a high-confidence source for subsequent frames.
This yields a \textit{temporal chain reaction} where recovered objects can propagate confidence to later frames, maintaining trajectory continuity beyond a single frame pair.
In this stage, the proposed class-conditional propagation mechanism ensures that confidence is transferred only between semantically consistent detections, maintaining high precision while improving recall across diverse object types.


\subsection{Implementation Details} 
\label{sec:imp_details}

Our model employs the same backbone architecture as~\cite{li2024slack, li2023ovtrack}, utilizing a Faster R-CNN detector with ResNet-50~\cite{he2016deep}. Following OVTrack's training protocol, the detector is trained on base classes from the LVIS dataset.
For the multi-cue fusion module (Section~\ref{sec:multi_cue}), the Self-attentive gated network (SGN) is designed as two fully connected layers with a ReLU activation.
Similarly, the Multi-cue fusion network (MFN) also consists of two fully connected layers with a ReLU activation.
For MGA (Section~\ref{sec:MGA}), we implement a single-head cross-attention mechanism with learnable projection matrices $\mathbf{W}_q, \mathbf{W}_k, \mathbf{W}_v \in \mathbb{R}^{d \times d}$, where the feature dimension $d = 256$.

In Section~\ref{sec:multi_cue}, for inter-frame confidence estimation, we compute the adaptive temperature parameter by $\alpha = \frac{\log\left(\frac{\delta}{1-\delta}\cdot \max(n,m)\right)}{\epsilon}$ with the temperature scaling parameters $\delta = 0.5$ and $\epsilon = 0.1$.
In Section~\ref{sec:MGA}, we use IoC threshold $\tau_{\text{ioc}} = 0.8$ for identifying parent-child relationships, TAQ threshold $\tau_q = 0.3$ for filtering unreliable children, and enhancement ratio $\lambda = 0.1$ for the residual connection.
In Section~\ref{sec:tcp}, we set the high-confidence threshold $\tau_{\text{high}} = 0.5$ and low-confidence threshold $\tau_{\text{low}} = 0.05$ for confidence thresholding.
The weight in Eq.~(\ref{eq:w}) is set as $\beta = 0.3$.
With respect to the value assignment of $\eta_j$ in Eq.~(\ref{eq:sj}), it is empirically set as $0.7$ under strong evidence ($\bar{w}_j$ and $|\mathcal{N}_j|$ exceed 0.8 and 3, respectively), as $0.5$ under moderate evidence ($\bar{w}_j$ and $|\mathcal{N}_j|$ exceed 0.6 and 2, respectively), as $0.3$ under weak evidence ($\bar{w}_j$ and $|\mathcal{N}_j|$ exceed 0.5 and 1, respectively).
We select the top 50 candidates with the highest scores in $[\tau_{\text{low}}, \tau_{\text{high}}]$ to limit computational cost.

For association training, we use the strict base-only C-TAO training annotations under the OVMOT protocol.
The model is trained for 10 epochs on 4 RTX 3090 GPUs, with MGA operating as a feature enhancement module during training (Section~\ref{sec:MGA}). 

For inference, we employ class-agnostic NMS for object filtering, with a maximum of 80 detected objects per frame. The final association feature ${\textbf{f}}_{\text{asso}}^{i}$ is used for object association with the bi-softmax matching strategy as~\cite{li2023ovtrack}. The matching process uses a threshold of $0.35$ and a memory queue length of $30$. Additionally, TCP (Section~\ref{sec:tcp}) is applied as an online post-processing module to recover low-confidence detections before final association.

\section{Experiments}
\label{sec:experiment}

\subsection{Datasets and Metrics}

Following~\cite{li2023ovtrack, Li_2024_CVPR, li2024slack, li2025open, li2025attention, li2025ovtr}, we conduct our evaluation using the same dataset and metrics. Specifically, we utilize the dataset TAO, which shares a similar category division scheme with LVIS~\cite{gupta2019lvis} for OVMOT evaluation. We designate the rare categories in LVIS as novel classes, while the remaining categories serve as base classes. 
Comparative experiments are conducted on the validation and test sets of TAO.
For performance evaluation, we adopt the standard OVMOT metric, tracking-everything accuracy (TETA)~\cite{li2022tracking}, which evaluates localization accuracy (LocA), classification accuracy (ClsA), and association accuracy (AssocA). To comprehensively demonstrate our algorithm's performance, we evaluate base and novel classes separately.

\revised{For the dense C-TAO-val benchmark, we additionally use Temporal continuity accuracy (TCA) as a class-agnostic adjacent-frame continuity diagnostic. TCA is computed with one-to-one class-agnostic matching at IoU 0.5. A ground-truth temporal link connects two visible boxes from the same trajectory in consecutive frames, and a predicted temporal link is induced by the same predicted track ID in consecutive frames. A predicted link is a temporal true positive only if both endpoints match the same ground-truth trajectory. We compute
\[
\mathrm{TCA}=
\frac{\mathrm{TC}_{\mathrm{TP}}}
{\mathrm{TC}_{\mathrm{TP}}+\mathrm{TC}_{\mathrm{FN}}+\mathrm{TC}_{\mathrm{FP}}}.
\]
Here, \(\mathrm{TC}_{\mathrm{FN}}=|\mathcal{L}_{\mathrm{GT}}|-\mathrm{TC}_{\mathrm{TP}}\), and \(\mathrm{TC}_{\mathrm{FP}}=|\mathcal{L}_{\mathrm{Pred}}^{\mathrm{rel}}|-\mathrm{TC}_{\mathrm{TP}}\), where \(\mathcal{L}_{\mathrm{Pred}}^{\mathrm{rel}}\) keeps target-related predicted links with at least one endpoint matched to a ground-truth box. Links whose two endpoints are both unmatched are excluded because TCA is designed to diagnose target-related temporal continuity; target-independent false positives are already reflected by TETA/LocA. TCA is therefore not intended to replace TETA or AssocA, but should be interpreted together with TETA/LocA/AssocA/ClsA.} 

\subsection{Comparisons with State-of-the-Art Methods}

We compare our method with current mainstream and state-of-the-art tracking methods on both the validation and test sets of TAO.
For a fair comparison, all methods utilize ResNet-50 as the backbone architecture. 
The comparison includes closed-set baselines trained on all categories, established off-the-shelf trackers such as ByteTrack~\cite{zhang2022bytetrack}, OC-SORT~\cite{cao2023observation}, and MASA~\cite{Li_2024_CVPR}, as well as specialized OVMOT methods like OVTrack~\cite{li2023ovtrack}, SLAck~\cite{li2024slack}, OVSORT~\cite{li2025open}, and TRACT~\cite{li2025attention}.

\begin{table*}[!t]
	\renewcommand{\arraystretch}{0.94}
	\centering
	\fontsize{7.7pt}{8.3pt}\selectfont
	\caption{Comparison of performance on validation and test sets. We compare the methods on open-vocabulary TAO benchmark~\cite{li2023ovtrack}. All methods use the same backbone. \textdagger~represents using the same detector. \revised{\textsuperscript{C} denotes reproduced results with C-TAO training.} (\%)}
	\label{tab:cmp}
	\begin{tabular}{l|l|cc|cccc|cccc}
		\hline
		\multirow{2}{*}{\textbf{Method}} & \multirow{2}{*}{\textbf{Venue (year)}} & \multicolumn{2}{c|}{\textbf{Classes}} & \multicolumn{4}{c|}{\textbf{Novel}} & \multicolumn{4}{c}{\textbf{Base}} \\ \cline{3-12}
		& & Novel & Base & TETA & LocA & AssocA & ClsA & TETA & LocA & AssocA & ClsA \\
		\hline
		\multicolumn{12}{l}{\textbf{Validation set}} \\
		\hline
		QDTrack~\cite{fischer2023qdtrack} & TPAMI (2023) & \checkmark & \checkmark & 22.5 & 42.7 & 24.4 & 0.4 & 27.1 & 45.6 & 24.7 & 11.0 \\
		\revised{QDTrack\textsuperscript{C}~\cite{fischer2023qdtrack}} & \revised{TPAMI (2023)} & \revised{\checkmark} & \revised{\checkmark} & \revised{23.2} & \revised{43.1} & \revised{26.1} & \revised{0.5} & \revised{28.0} & \revised{46.0} & \revised{26.5} & \revised{11.6} \\
		TETer~\cite{li2022tracking} & ECCV (2022) & \checkmark & \checkmark & 25.7 & 45.9 & 31.1 & 0.2 & 30.3 & 47.4 & 31.6 & 12.1 \\
		\revised{TETer\textsuperscript{C}~\cite{li2022tracking}} & \revised{ECCV (2022)} & \revised{\checkmark} & \revised{\checkmark} & \revised{26.7} & \revised{46.4} & \revised{33.4} & \revised{0.3} & \revised{31.4} & \revised{47.9} & \revised{33.2} & \revised{13.1} \\
		DeepSORT (ViLD)~\cite{wojke2017simple} & ICIP (2017) & - & \checkmark & 21.1 & 46.4 & 14.7 & 2.3 & 26.9 & 47.1 & 15.8 & 17.7 \\
		Tracktor++ (ViLD)~\cite{bergmann2019tracking} & ICCV (2019) & - & \checkmark & 22.7 & 46.7 & 19.3 & 2.2 & 28.3 & 47.4 & 20.5 & 17.0 \\
		ByteTrack\textsuperscript{\textdagger}~\cite{zhang2022bytetrack} & ECCV (2022) & - & \checkmark & 22.0 & 48.2 & 16.6 & 1.0 & 28.2 & 50.4 & 18.1 & 16.0 \\
		OC-SORT\textsuperscript{\textdagger}~\cite{cao2023observation} & CVPR (2023) & - & \checkmark & 23.7 & 49.6 & 20.4 & 1.1 & 28.9 & 51.4 & 19.8 & 15.4 \\
		OVTrack\textsuperscript{\textdagger}~\cite{li2023ovtrack} & CVPR (2023) & - & \checkmark & 27.8 & 48.8 & 33.6 & 1.5 & 35.5 & 49.3 & 36.9 & 20.2 \\
		\revised{OVTrack\textsuperscript{\textdagger,C}~\cite{li2023ovtrack}} & \revised{CVPR (2023)} & \revised{-} & \revised{\checkmark} & \revised{30.5} & \revised{50.2} & \revised{39.3} & \revised{2.1} & \revised{37.4} & \revised{52.7} & \revised{39.3} & \revised{20.1} \\
		MASA (R50)\textsuperscript{\textdagger}~\cite{Li_2024_CVPR} & CVPR (2024) & - & - & 30.0 & 54.2 & 34.6 & 1.0 & 36.9 & 55.1 & 36.4 & 19.3 \\
		SLAck\textsuperscript{\textdagger}~\cite{li2024slack} & ECCV (2024) & - & \checkmark & 31.1 & 54.3 & 37.8 & 1.3 & 37.2 & 55.0 & 37.6 & 19.1 \\
		OVSORT\textsuperscript{\textdagger}~\cite{li2025open} & TMM (2025) & - & \checkmark & 30.8 & 53.0 & 37.6 & 1.9 & 38.2 & 55.3 & 39.9 & 19.4 \\
		TRACT\textsuperscript{\textdagger}~\cite{li2025attention} & ICCV (2025) & - & \checkmark & 31.3 & 52.7 & 37.8 & 3.4 & 38.5 & 55.0 & 39.0 & \textbf{21.5} \\
		OVTR\textsuperscript{\textdagger}~\cite{li2025ovtr} & ICLR (2025) & - & \checkmark & 31.4 & 54.4 & 34.5 & \textbf{5.4} & 36.6 & 52.2 & 37.6 & 20.1 \\
		\revised{OVTR\textsuperscript{\textdagger,C}~\cite{li2025ovtr}} & \revised{ICLR (2025)} & \revised{-} & \revised{\checkmark} & \revised{32.5} & \revised{54.5} & \revised{37.6} & \revised{5.3} & \revised{37.2} & \revised{52.5} & \revised{38.9} & \revised{20.1} \\
		\revised{COVTrack\textsuperscript{\textdagger,C}~\cite{qian2025covtrack}} & \revised{ICCV (2025)} & \revised{-} & \revised{\checkmark} & \revised{34.3} & \revised{58.2} & \revised{41.3} & \revised{3.5} & \revised{39.6} & \revised{57.3} & \revised{42.0} & \revised{19.6} \\
		\revised{COVTrack++\textsuperscript{\textdagger,C}} & \revised{-} & \revised{-} & \revised{\checkmark} & \revised{\textbf{35.4}} & \revised{\textbf{60.1}} & \revised{\textbf{42.6}} & \revised{3.5} & \revised{\textbf{40.3}} & \revised{\textbf{58.6}} & \revised{\textbf{42.6}} & \revised{19.6} \\
		\hline
		\multicolumn{12}{l}{\textbf{Test set}} \\
		\hline
		QDTrack~\cite{fischer2023qdtrack} & TPAMI (2023) & \checkmark & \checkmark & 20.2 & 39.7 & 20.9 & 0.2 & 25.8 & 43.2 & 23.5 & 10.6 \\
		\revised{QDTrack\textsuperscript{C}~\cite{fischer2023qdtrack}} & \revised{TPAMI (2023)} & \revised{\checkmark} & \revised{\checkmark} & \revised{20.8} & \revised{40.2} & \revised{22.0} & \revised{0.3} & \revised{26.4} & \revised{43.6} & \revised{24.7} & \revised{10.9} \\
		TETer~\cite{li2022tracking} & ECCV (2022) & \checkmark & \checkmark & 21.7 & 39.1 & 25.9 & 0.0 & 28.4 & 44.0 & 30.4 & 10.7 \\
		\revised{TETer\textsuperscript{C}~\cite{li2022tracking}} & \revised{ECCV (2022)} & \revised{\checkmark} & \revised{\checkmark} & \revised{22.5} & \revised{39.8} & \revised{27.6} & \revised{0.1} & \revised{28.9} & \revised{44.4} & \revised{31.5} & \revised{10.8} \\
		DeepSORT (ViLD)~\cite{wojke2017simple} & ICIP (2017) & - & \checkmark & 17.2 & 38.4 & 11.6 & 1.7 & 24.5 & 43.3 & 14.6 & 15.2 \\
		Tracktor++ (ViLD)~\cite{bergmann2019tracking} & ICCV (2019) & - & \checkmark & 18.0 & 39.0 & 13.4 & 1.7 & 26.0 & 44.1 & 19.0 & 14.8 \\
		OVTrack\textsuperscript{\textdagger}~\cite{li2023ovtrack} & CVPR (2023) & - & \checkmark & 24.1 & 41.8 & 28.7 & 1.8 & 32.6 & 45.6 & 35.4 & 16.9 \\
		\revised{OVTrack\textsuperscript{\textdagger,C}~\cite{li2023ovtrack}} & \revised{CVPR (2023)} & \revised{-} & \revised{\checkmark} & \revised{25.3} & \revised{42.3} & \revised{31.5} & \revised{2.0} & \revised{33.3} & \revised{46.6} & \revised{36.4} & \revised{16.8} \\
		SLAck\textsuperscript{\textdagger}~\cite{li2024slack} & ECCV (2024) & - & \checkmark & 27.1 & 49.1 & 30.0 & 2.0 & 34.7 & 52.5 & 35.6 & 16.1 \\
		OVSORT\textsuperscript{\textdagger}~\cite{li2025open} & TMM (2025) & - & \checkmark & 28.1 & 48.0 & 33.4 & 2.7 & 35.1 & 51.6 & 38.3 & 15.4 \\
		TRACT\textsuperscript{\textdagger}~\cite{li2025attention} & ICCV (2025) & - & \checkmark & 27.3 & 48.2 & 30.7 & 3.1 & 36.2 & 52.3 & 39.1 & \textbf{17.2} \\
		OVTR\textsuperscript{\textdagger}~\cite{li2025ovtr} & ICLR (2025) & - & \checkmark & 27.1 & 47.1 & 32.1 & 2.1 & 34.5 & 51.1 & 37.5 & 14.9 \\
		\revised{OVTR\textsuperscript{\textdagger,C}~\cite{li2025ovtr}} & \revised{ICLR (2025)} & \revised{-} & \revised{\checkmark} & \revised{27.6} & \revised{47.4} & \revised{33.4} & \revised{2.0} & \revised{34.9} & \revised{51.6} & \revised{38.0} & \revised{15.0} \\
		\revised{COVTrack\textsuperscript{\textdagger,C}~\cite{qian2025covtrack}} & \revised{ICCV (2025)} & \revised{-} & \revised{\checkmark} & \revised{28.9} & \revised{50.9} & \revised{32.6} & \revised{\textbf{3.3}} & \revised{37.9} & \revised{54.5} & \revised{42.1} & \revised{\textbf{17.2}} \\
		\revised{COVTrack++\textsuperscript{\textdagger,C}} & \revised{-} & \revised{-} & \revised{\checkmark} & \revised{\textbf{30.5}} & \revised{\textbf{52.5}} & \revised{\textbf{35.8}} & \revised{3.2} & \revised{\textbf{38.9}} & \revised{\textbf{57.4}} & \revised{\textbf{42.3}} & \revised{17.0} \\
		\hline
	\end{tabular}
\end{table*}

\revised{Table~\ref{tab:cmp} reports the original published results together with reproduced C-TAO training results for methods with available public code/training artifacts, including all reproducible public-code OVMOT methods in the table. Across both the original published setting and the matched C-TAO training setting, COVTrack++ achieves clearly better results than existing methods on validation and test sets. Under C-TAO training, it further improves over COVTrack by +1.1/+1.6 novel TETA and +1.3/+3.2 novel AssocA on validation/test, showing a clear gain especially on the more challenging test split.}   

\subsection{\revised{Dense C-TAO-val Benchmark}}
\label{sec:ctao_dense_benchmark}

\revised{The standard sparse TAO validation/test protocol is kept as the primary benchmark for comparison with prior work. To further evaluate tracking behavior with continuous validation annotations, we report a dense C-TAO-val benchmark. C-TAO-val provides frame-dense annotations for TAO validation trajectories, enabling more fine-grained evaluation of adjacent-frame continuity than sparse TAO-val. All methods in Table~\ref{tab:ctao_val_benchmark} are trained on C-TAO-train and evaluated on C-TAO-val: closed-set reference rows use base+novel C-TAO-train labels, while OVMOT rows use the strict base-only C-TAO-train file. We include as many publicly reproducible methods as possible, including closed-set trackers (QDTrack and TETer) and OVMOT methods (OVTrack, OVTR, COVTrack, and COVTrack++); other methods are not listed in this dense table because their code is not publicly available. Overall, COVTrack++ achieves the best dense C-TAO-val TCA and TETA/LocA/AssocA, confirming improved tracking accuracy and adjacent-frame temporal continuity.}   

\begin{table*}[!t]
	\renewcommand{\arraystretch}{1.05}
	\scriptsize
	\setlength{\tabcolsep}{3.6pt}
	\centering
	\caption{\revised{Dense C-TAO-val benchmark. All methods use the same backbone. (\%)}}
	\label{tab:ctao_val_benchmark}
	\begin{tabular}{l|cc|cccc|cccc|c}
		\hline
		\multirow{2}{*}{\textbf{Method}} & \multicolumn{2}{c|}{\textbf{Classes}} & \multicolumn{4}{c|}{\textbf{Novel}} & \multicolumn{4}{c|}{\textbf{Base}} & \multirow{2}{*}{\textbf{TCA}} \\ \cline{2-11}
		& Novel & Base & TETA & LocA & AssocA & ClsA & TETA & LocA & AssocA & ClsA &  \\
		\hline
		\revised{QDTrack~\cite{fischer2023qdtrack}} & \revised{\checkmark} & \revised{\checkmark} & \revised{20.9} & \revised{41.5} & \revised{20.8} & \revised{0.4} & \revised{25.3} & \revised{43.8} & \revised{21.4} & \revised{10.7} & \revised{22.6} \\
		\revised{TETer~\cite{li2022tracking}} & \revised{\checkmark} & \revised{\checkmark} & \revised{24.1} & \revised{44.9} & \revised{27.1} & \revised{0.3} & \revised{28.6} & \revised{46.2} & \revised{27.5} & \revised{12.1} & \revised{27.8} \\
		\hline
		\revised{OVTrack~\cite{li2023ovtrack}} & \revised{-} & \revised{\checkmark} & \revised{28.7} & \revised{49.6} & \revised{34.5} & \revised{2.0} & \revised{35.3} & \revised{52.0} & \revised{34.0} & \revised{\textbf{19.9}} & \revised{34.2} \\
		\revised{OVTR~\cite{li2025ovtr}} & \revised{-} & \revised{\checkmark} & \revised{30.6} & \revised{53.4} & \revised{34.1} & \revised{\textbf{4.3}} & \revised{35.4} & \revised{53.9} & \revised{34.0} & \revised{18.3} & \revised{36.4} \\
		\revised{COVTrack~\cite{qian2025covtrack}} & \revised{-} & \revised{\checkmark} & \revised{33.0} & \revised{57.3} & \revised{37.6} & \revised{4.1} & \revised{38.4} & \revised{57.4} & \revised{38.6} & \revised{19.1} & \revised{38.9} \\
		\revised{COVTrack++} & \revised{-} & \revised{\checkmark} & \revised{\textbf{34.6}} & \revised{\textbf{59.9}} & \revised{\textbf{39.7}} & \revised{\textbf{4.3}} & \revised{\textbf{39.7}} & \revised{\textbf{59.7}} & \revised{\textbf{40.3}} & \revised{19.2} & \revised{\textbf{40.5}} \\
		\hline
	\end{tabular}
\end{table*}

\subsection{\revised{Attribution of Supervision and Algorithmic Gains}}
\label{sec:attribution_analysis}

\revised{Table~\ref{tab:attribution_control} separates supervision and method gains. C-TAO consistently improves all variants by +3.2 to +3.6 novel TETA. Under fixed training supervision, COVTrack (MCF only) improves over the module-free baseline by +3.6/+3.7, and COVTrack++ increases the total method gain to +4.4/+4.8. The fixed-C-TAO comparison further shows that COVTrack++ improves over COVTrack by +1.1 novel TETA and +1.3 novel AssocA on TAO validation, with larger gains on the more challenging TAO test split (+1.6 novel TETA and +3.2 novel AssocA), confirming the effectiveness of the journal-extension modules.}

\begin{table}[!t]
	\renewcommand{\arraystretch}{1.05}
	\scriptsize
	\setlength{\tabcolsep}{2.4pt}
	\centering
	\caption{\revised{Attribution control for separating C-TAO supervision gains and fixed-supervision algorithmic gains on TAO. (\%)}}
	\label{tab:attribution_control}
	\vspace{-8pt}
	\revised{\textbf{TAO validation novel TETA attribution matrix}}\\[2pt]
		\begin{tabular}{l|cc|c}
			\hline
			\revised{\textbf{Method}} & \revised{\textbf{Orig. TAO}} & \revised{\textbf{C-TAO}} & \revised{\textbf{C-TAO gain}} \\
			\hline
				\revised{Module-free (no MCF/MGA/TCP)} & \revised{27.4} & \revised{30.6} & \revised{+3.2} \\
				\revised{COVTrack (MCF only)} & \revised{31.0} & \revised{34.3} & \revised{+3.3} \\
				\revised{COVTrack++ (full)} & \revised{31.8} & \revised{35.4} & \revised{+3.6} \\
				\hline
				\revised{COVTrack gain over module-free} & \revised{+3.6} & \revised{+3.7} & \revised{--} \\
				\revised{COVTrack++ gain over module-free} & \revised{+4.4} & \revised{+4.8} & \revised{--} \\
			\hline
		\end{tabular}
	\vspace{5pt}

	\revised{\textbf{Fixed C-TAO extension comparison}}\\[2pt]
		\begin{tabular}{l|cc|cc}
			\hline
			\multirow{2}{*}{\revised{\textbf{Split}}} & \multicolumn{2}{c|}{\revised{\textbf{Novel TETA}}} & \multicolumn{2}{c}{\revised{\textbf{Novel AssocA}}} \\
			& \revised{COVTrack \(\rightarrow\) COVTrack++} & \revised{Gain} & \revised{COVTrack \(\rightarrow\) COVTrack++} & \revised{Gain} \\
			\hline
			\revised{TAO val} & \revised{34.3 \(\rightarrow\) 35.4} & \revised{+1.1} & \revised{41.3 \(\rightarrow\) 42.6} & \revised{+1.3} \\
			\revised{TAO test} & \revised{28.9 \(\rightarrow\) 30.5} & \revised{+1.6} & \revised{32.6 \(\rightarrow\) 35.8} & \revised{+3.2} \\
			\hline
		\end{tabular}
	\vspace{-6pt}
\end{table}

\subsection{Ablation Study}
\label{sec:ablation}

We conduct comprehensive ablation studies to validate the effectiveness of our proposed framework on the TAO validation set.
\revised{Table~\ref{tab:main_ablation} reports framework-level module ablations. We include w/o MGA\&TCP as the architecture-level regression point to COVTrack under the current evaluation setting, while w/o TCP keeps MGA and isolates the effect of MGA without TCP.}

\begin{table}[!htbp]
	\renewcommand{\arraystretch}{1.05}
	\scriptsize
	\setlength{\tabcolsep}{2.0pt}
	\centering
	\caption{\revised{Module-level ablation results on the TAO validation set. (\%)}} 
	\label{tab:main_ablation}
	\resizebox{\columnwidth}{!}{%
	\begin{tabular}{l|cccc|cccc}
		\hline
		\multirow{2}{*}{\textbf{Method}} & \multicolumn{4}{c|}{\textbf{Novel}} & \multicolumn{4}{c}{\textbf{Base}} \\ \cline{2-9}
		& TETA & LocA & AssocA & ClsA & TETA & LocA & AssocA & ClsA \\
		\hline
		\revised{w/o MGA\&TCP (COVTrack)} & \revised{34.3} & \revised{58.2} & \revised{41.3} & \revised{\textbf{3.5}} & \revised{39.6} & \revised{57.3} & \revised{42.0} & \revised{\textbf{19.6}} \\
		w/o MCF & 33.9 & 59.0 & 40.5 & 2.1 & 39.2 & 58.0 & 40.2 & 19.4 \\
		w/o MGA & 34.6 & 59.2 & 41.5 & 3.2 & 40.0 & 58.3 & 42.2 & 19.5 \\
		\revised{w/o TCP} & \revised{34.6} & \revised{58.4} & \revised{42.0} & \revised{3.3} & \revised{39.8} & \revised{57.4} & \revised{42.5} & \revised{19.4} \\
		Ours (Full) & \textbf{35.4} & \textbf{60.1} & \textbf{42.6} & \textbf{3.5} & \textbf{40.3} & \textbf{58.6} & \textbf{42.6} & \textbf{19.6} \\
		\hline
	\end{tabular}}
\end{table}

\revised{The results show that each module contributes to the final performance. Removing MCF mainly reduces AssocA, confirming the value of confidence-guided multi-cue fusion for association. Removing MGA also decreases novel AssocA from 42.6\% to 41.5\%, showing the benefit of reliable multi-granularity evidence. Removing TCP lowers novel LocA from 60.1\% to 58.4\%, verifying its role in recovering low-confidence detections. Compared with the COVTrack regression baseline, adding MGA without TCP improves AssocA from 41.3\% to 42.0\% on novel categories and from 42.0\% to 42.5\% on base categories, while LocA and ClsA remain close. The full model achieves the best TETA, LocA, and AssocA on both novel and base categories.}

\subsection{In-Depth Analysis of Each Module}
\label{sec:depth}
Moreover, we perform isolated analysis of each module to verify the effectiveness of detailed strategies within each module.

\subsubsection{In-Depth Analysis of MCF Module} 
To understand the internal mechanisms of the adaptive fusion module, we conduct a detailed ablation study focusing exclusively on the confidence-based multi-cue fusion strategy, with results presented in Table~\ref{tab:ablation}.
This analysis isolates the adaptive fusion module (equivalent to using only this module without MGA and TCP) to systematically evaluate each confidence component and feature contribution.

\begin{table}[!htbp]
	\renewcommand{\arraystretch}{1.2}
	\scriptsize
	\setlength{\tabcolsep}{1.5pt}
	\centering
	\caption{Component-level ablation of the adaptive multi-cue fusion module on the TAO validation set (without MGA and TCP). (\%)} 
	\label{tab:ablation}
	\resizebox{\columnwidth}{!}{%
	\begin{tabular}{l|cccc|cccc}
		\hline
		\multirow{2}{*}{\textbf{Ablation Method}} & \multicolumn{4}{c|}{\textbf{Novel}} & \multicolumn{4}{c}{\textbf{Base}} \\ \cline{2-9}
		& TETA & LocA & AssocA & ClsA & TETA & LocA & AssocA & ClsA \\
		\hline
		\ding{192} w/o $c^{\text{intra}}_{\text{sem}}$ & 33.3 & 57.3 & 40.6 & 2.1 & 39.1 & 57.3 & 41.2 & 18.9 \\
		\ding{193} w/o $c^{\text{intra}}_{\text{loc}}$ & 32.8 & 57.6 & 37.7 & 3.1 & 38.7 & 57.4 & 39.5 & 19.1 \\
		\ding{194} w/o $c^{\text{inter}}_{\text{sem}}$ & 33.1 & 57.8 & 39.1 & 2.3 & 38.6 & 57.6 & 39.6 & 18.6 \\
		\ding{195} w/o $c^{\text{inter}}_{\text{loc}}$ & 32.8 & 57.8 & 37.3 & 3.2 & 38.2 & 57.4 & 38.4 & 18.8 \\
		\ding{196} w/o $c^{\text{inter}}_{\text{app}}$ & 33.4 & 57.8 & 39.2 & 3.3 & 38.6 & 57.4 & 39.2 & 19.1 \\ \hline
		\ding{197} w/o $c^{\text{intra}}_{\text{sem}}$+$c^{\text{intra}}_{\text{loc}}$ (intra conf.) & 32.3 & 57.7 & 36.8 & 2.3 & 38.5 & 57.4 & 39.3 & 18.8 \\
		\ding{198} w/o $c^{\text{inter}}_{\text{sem}}$+$c^{\text{inter}}_{\text{loc}}$ (inter conf.) & 32.1 & 57.7 & 35.7 & 2.8 & 37.8 & 57.7 & 37.3 & 18.3 \\ \hline
		\ding{199} w/o $\mathbf{e}_{\text{app}}$ in Eq.~\myeqref{eq:SGN}  & 32.0 & 57.4 & 36.5 & 2.2 & 38.4 & 57.3 & 39.2 & 18.7 \\
		\ding{200} w/o $\tilde{\mathbf{e}}_{\text{sem}}$ in Eq.~\myeqref{eq:m-cue}  & 33.0 & 57.0 & 39.7 & 2.4 & 38.2 & 57.1 & 40.1 & 17.5 \\
		\ding{201} w/o $\tilde{\mathbf{e}}_{\text{loc}}$ in Eq.~\myeqref{eq:m-cue} & 33.0 & 57.5 & 38.4 & 3.2 & 38.2 & 57.1 & 39.2 & 18.5 \\
		\circled{11} w/o $\textbf{e}_{\text{app}}$ in Eq.~\myeqref{eq:refine}  & 33.3 & 57.8 & 39.1 & 2.9 & 38.6 & 57.6 & 39.4 & 18.7 \\ \hline
		Full MCF & \textbf{34.3} & \textbf{58.2} & \textbf{41.3} & \textbf{3.5} & \textbf{39.6} & \textbf{57.3} & \textbf{42.0} & \textbf{19.6} \\
		\hline
	\end{tabular}}
\end{table}

\textit{\ding{202} Confidence mechanisms.}
The ablation results demonstrate that the proposed cue confidence mechanisms effectively enhance association performance.
\revised{The motion-related cues have a larger impact than semantic-related cues, indicating the importance of continuous TAO supervision for learning temporal continuity features.}
\revised{Furthermore, the intra-frame and inter-frame cue confidence mechanisms jointly improve the discriminative ability of the fused features.}
\revised{Table~\ref{tab:cycle_confidence} further compares the self-product diagonal baseline with our forward-backward cycle confidence. Replacing reciprocal cycle confidence with self-product diagonal confidence decreases novel TETA from 34.3\% to 33.4\% and novel AssocA from 41.3\% to 39.2\%, confirming that reciprocal temporal consistency provides reliability beyond one-way distribution sharpness.}   

\begin{table}[!htbp]
	\renewcommand{\arraystretch}{1.05}
	\scriptsize
	\setlength{\tabcolsep}{1.8pt}
	\centering
	\caption{\revised{Controlled ablation of inter-frame confidence estimation under the MCF-only setting. (\%)}}
	\label{tab:cycle_confidence}
	\resizebox{\columnwidth}{!}{%
	\begin{tabular}{l|cccc|cccc}
		\hline
		\multirow{2}{*}{\textbf{Inter-frame confidence}} & \multicolumn{4}{c|}{\textbf{Novel}} & \multicolumn{4}{c}{\textbf{Base}} \\ \cline{2-9}
		& TETA & LocA & AssocA & ClsA & TETA & LocA & AssocA & ClsA \\
		\hline
		\revised{Self-product diagonal confidence} & \revised{33.4} & \revised{57.8} & \revised{39.2} & \revised{3.2} & \revised{38.8} & \revised{57.1} & \revised{40.1} & \revised{19.3} \\
		\revised{Forward-backward cycle confidence (ours)} & \revised{\textbf{34.3}} & \revised{\textbf{58.2}} & \revised{\textbf{41.3}} & \revised{\textbf{3.5}} & \revised{\textbf{39.6}} & \revised{\textbf{57.3}} & \revised{\textbf{42.0}} & \revised{\textbf{19.6}} \\
		\hline
	\end{tabular}}
\end{table}

\textit{\ding{203} Feature fusion effectiveness.}
We then analyze the effectiveness of feature fusion.
The ablation results validate the crucial role of appearance features as input during the intra-frame confidence learning process in SGN. 
The absence of appearance features for object representations leads to a substantial decrease in confidence estimation effectiveness, particularly for novel classes.
Additionally, the results confirm that both semantic and location features contribute to the performance improvement, with location features showing a more important impact on the association.
The refinement operation can effectively improve association results.

\textit{\ding{204} Multi-cue feature fusion comparison.}
To further validate the superiority of our confidence-based adaptive fusion over naive feature combination strategies, we compare with SLAck~\cite{li2024slack}, which employs simple summation for multi-cue fusion. 
SLAck is the only existing OVMOT method that trains on TAO videos and uses multi-cue fusion (appearance, motion, semantic), making it the most relevant baseline.
For fair comparison, we use only the Multi-cue adaptive fusion (MCF) module (without MGA and TCP), maintaining the same detector setup as SLAck.

\begin{table*}[!t]
	\renewcommand{\arraystretch}{1}
	\footnotesize
	\centering
	\caption{Comparison of fusion strategies and training data quality. SLAck uses simple summation for multi-cue fusion, while Ours$^*$ uses adaptive fusion. Both methods use the same detector for fair comparison. (\%)}
	\label{tab:slack_comparison}
	\begin{tabular}{l|cccc|cccc}
		\hline
		\multirow{2}{*}{\textbf{Method}} & \multicolumn{4}{c|}{\textbf{Novel}} & \multicolumn{4}{c}{\textbf{Base}} \\ \cline{2-9}
		& TETA & LocA & AssocA & ClsA & TETA & LocA & AssocA & ClsA \\
		\hline
		SLAck trained on (original) TAO & 24.6 & 47.1 & 24.1 & 2.5 & - & - & - & - \\
		Ours$^*$ trained on (original) TAO & 31.0 & 55.5 & \underline{34.8} (+10.7) & 2.8 & 36.3 & 54.4 & \underline{35.8} & 18.7 \\ \hline
		SLAck trained on (original) TAO w pseudo labels & 31.1 & 54.3 & 37.8 & 1.3 & 37.2 & 55.0 & 37.6 & 19.1 \\
		Ours$^*$ trained on (original) TAO w pseudo labels & 32.8 & 56.4 & 39.2  & 2.9 & 38.3 & 56.6 & 38.9  & 19.3 \\ \hline
		Ours$^*$ w C-TAO & {34.3} & {58.2} & {41.3} (+6.5) & {3.5} & {39.6} & {57.3} & {42.0} (+6.2) & {19.6} \\
		\hline
		\multicolumn{9}{l}{Ours$^*$ denotes our method using only MCF module without MGA \& TCP.} \\
	\end{tabular}
\end{table*}

As shown in Table~\ref{tab:slack_comparison}, we first compare under sparse and pseudo-labeled supervision to isolate the effect of fusion strategies.
When trained on the original sparse TAO annotations, SLAck's performance degrades significantly (novel TETA: 24.6\%), as equal-weight summation amplifies noise from unreliable motion and semantic features under sparse supervision.
In contrast, our adaptive fusion maintains substantially better performance (novel TETA: 31.0\%, +6.4\%), achieving a remarkable 10.7\% improvement in novel AssocA (34.8\% vs 24.1\%).
This demonstrates that confidence-based adaptive weighting is crucial for handling unreliable cues, dynamically down-weighting unstable features while amplifying stable ones.

When trained on TAO with pseudo labels (generated via IoU matching), both methods improve due to increased temporal continuity, but our adaptive fusion still outperforms SLAck by +1.7\% in novel TETA (32.8\% vs 31.1\%) and +1.4\% in novel AssocA (39.2\% vs 37.8\%).
This validates that adaptive fusion generalizes better than simple summation even with continuous supervision, as it dynamically adjusts cue weights based on their varying reliability across different scenarios and object categories.
These comparisons confirm the effectiveness of our confidence-based fusion strategy regardless of annotation quality.

\subsubsection{In-Depth Analysis of MGA Module}
We further validate key design choices within MGA in Table~\ref{tab:module_ablation} (top two rows). 
\begin{table}[H]
	\renewcommand{\arraystretch}{1.03}
	\scriptsize
	\setlength{\tabcolsep}{1.8pt}
	\centering
	\caption{Component-level ablation of MGA and TCP modules on the TAO validation set. (\%)} 
	\label{tab:module_ablation}
	\resizebox{\columnwidth}{!}{%
	\begin{tabular}{l|cccc|cccc}
		\hline
		\multirow{2}{*}{\textbf{Method}} & \multicolumn{4}{c|}{\textbf{Novel}} & \multicolumn{4}{c}{\textbf{Base}} \\ \cline{2-9}
		& TETA & LocA & AssocA & ClsA & TETA & LocA & AssocA & ClsA \\
		\hline
		Replace MGA w/ average pooling & 34.3 & 59.0 & 40.9 & 3.1 & 39.6 & 58.1 & 41.3 & 19.3 \\
		w/o TAQ gating in MGA & 34.5 & 59.1 & 41.1 & 3.3 & 39.7 & 58.2 & 41.6 & 19.3 \\ \hline
		\revised{w/o class-conditional in TCP} & \revised{34.7} & \revised{58.8} & \revised{41.8} & \revised{3.4} & \revised{39.9} & \revised{57.9} & \revised{42.2} & \revised{19.6} \\
		\revised{w/o multi-cue edges in TCP} & \revised{34.3} & \revised{58.0} & \revised{41.6} & \revised{3.3} & \revised{39.5} & \revised{57.0} & \revised{42.1} & \revised{19.5} \\ \hline
		Ours (Full) & \textbf{35.4} & \textbf{60.1} & \textbf{42.6} & \textbf{3.5} & \textbf{40.3} & \textbf{58.6} & \textbf{42.6} & \textbf{19.6} \\
		\hline
	\end{tabular}}
\end{table}
Replacing learned cross-attention with simple average pooling degrades performance even further compared to completely removing MGA, with novel TETA dropping to 34.3\% (worse than w/o MGA's 34.6\%) and AssocA to 40.9\% (worse than 41.5\%).
This result demonstrates that naive aggregation of child features without adaptive weighting actually harms tracking performance: blindly incorporating all child information introduces noise from irrelevant or mismatched children, making the parent representation less discriminative than using no hierarchical information at all.
This validates the necessity of our learned cross-attention mechanism that selectively aggregates only relevant child features.
Similarly, removing TAQ gating also degrades performance below the w/o MGA baseline, with novel TETA dropping to 34.5\% and AssocA to 41.1\%.
This shows that without quality-aware filtering, unreliable or low-quality child features contaminate parent representations, leading to worse tracking results than simply not using hierarchical aggregation.
These results confirm that both components are essential for MGA to provide positive contributions: improper hierarchical aggregation is worse than no aggregation at all.

\subsubsection{In-Depth Analysis of TCP Module}
\revised{As shown in Table~\ref{tab:module_ablation} (the third and fourth rows), removing the class-conditional constraint only brings limited benefit over the w/o TCP baseline in Table~\ref{tab:main_ablation}: novel LocA increases from 58.4\% to 58.8\%, but AssocA decreases from 42.0\% to 41.8\%, leading to only 34.7\% novel TETA.
This indicates that unconstrained propagation can recover more candidates but also introduces spurious cross-category matches.
With the full class-conditional TCP, the model reaches 35.4\%/60.1\%/42.6\% in novel TETA/LocA/AssocA, confirming that semantic consistency is important for reliable confidence propagation.
Removing multi-cue edges further drops novel TETA/LocA/AssocA to 34.3\%/58.0\%/41.6\%, below the w/o TCP baseline, showing that appearance-only temporal matching is insufficient.
These results confirm that both class-conditional propagation and multi-cue temporal edges are needed for robust TCP.}

\subsection{Effectiveness of the Proposed C-TAO Dataset}

\begin{table*}[!t]
	\renewcommand{\arraystretch}{1}
	\footnotesize
	\centering
	\caption{Impact of annotation density on tracking performance. Comparison of OVTrack and our full framework trained on different annotation densities. (\%)}
	\label{tab:dataset_effectiveness}
	\begin{tabular}{c|l|cccc|cccc}
		\hline
		\multirow{2}{*}{\textbf{Method}} & \multirow{2}{*}{\textbf{Annotation ratio}} & \multicolumn{4}{c|}{\textbf{Novel}} & \multicolumn{4}{c}{\textbf{Base}} \\ \cline{3-10}
		&  & TETA & LocA & AssocA & ClsA & TETA & LocA & AssocA & ClsA \\
		\hline
		\multirow{3}{*}{OVTrack}
		& 3.7\% (original TAO)  & 25.4 & 48.8 & 25.6 & 1.9 & 33.5 & 49.7 & 30.9 & {19.8}\\
		& 3.7\% $\times$ 2 (P-TAO) & 29.5 & 50.2 & 36.5 & 1.7 & 36.5 & 51.7 & 38.2 & 19.6 \\
		& 100\% (C-TAO) & 30.5 & 50.2 & 39.3 (+13.7) & 2.1 & 37.4 & 52.7 & 39.3 (+8.4) & 20.1 \\
		\hline
		\multirow{3}{*}{Ours}
		& 3.7\% (original TAO) & 31.8 & 57.0 & 35.6 & 2.9 & 36.8 & 55.4 & 36.2 & 18.8 \\
		& 3.7\% $\times$ 2 (P-TAO) & 34.6 & 59.6 & 40.9 & 3.3 & 39.2 & 58.0 & 40.4 & 19.1 \\
		& 100\% (C-TAO)  & \textbf{35.4} & \textbf{60.1} & \textbf{42.6} & \textbf{3.5} & \textbf{40.3} & \textbf{58.6} & \textbf{42.6} & \textbf{19.6} \\
		\hline
	\end{tabular}
\end{table*}

\subsubsection{C-TAO vs Pseudo Labels}
While the comparisons in Table~\ref{tab:slack_comparison} demonstrate the superiority of our adaptive fusion strategy, they also reveal critical insights about training data quality for OVMOT.
As shown in Table~\ref{tab:slack_comparison}, when our adaptive fusion module is trained on C-TAO with ground-truth continuous annotations, it achieves 34.3\% novel TETA.
Compared to training on original sparse TAO (31.0\% novel TETA), C-TAO provides substantial improvements of +6.5\% and +6.2\% in novel and base AssocA, respectively.
Even compared to training on TAO with pseudo labels (32.8\% novel TETA), C-TAO still yields notable gains of +2.1\% and +3.1\% in novel and base AssocA, demonstrating the superiority of ground-truth continuous annotations over heuristic pseudo labeling.

This significant gain confirms that high-quality continuous annotations enable more effective learning of temporal dynamics compared to heuristic pseudo labeling via IoU matching.
Pseudo labels, while providing temporal continuity, suffer from propagation errors and lack temporal consistency, especially in dynamic scenarios with rapid motion or severe occlusions where IoU matching becomes unreliable.
In contrast, C-TAO's ground-truth annotations provide accurate supervision for smooth trajectories, gradual appearance changes, and intermediate states (e.g., partial occlusions), allowing the model to learn robust motion patterns and appearance-invariant features that generalize better to novel categories.
This validates the necessity and value of investing in high-quality continuous annotation for OVMOT training.

\subsubsection{Annotation Density Analysis}
To further understand the impact of annotation density on OVMOT training, we investigate how different levels of temporal continuity affect tracking performance.
The original TAO annotates only 3.7\% of frames, which proves insufficient for training effective trackers, as demonstrated by the degraded performance in Table~\ref{tab:slack_comparison}.

To understand the incremental benefits of increasing annotation density, we conduct experiments with Pairwise-TAO (P-TAO), which provides one continuous label after each original TAO frame, resulting in a 3.7\% $\times$ 2 annotation ratio.
We evaluate both OVTrack~\cite{li2023ovtrack} and our full framework across three annotation densities, with results shown in Table~\ref{tab:dataset_effectiveness}.
We can see that, the TAO$\rightarrow$P-TAO$\rightarrow$C-TAO progression reveals important insights about annotation density requirements for OVMOT.
For OVTrack, increasing from sparse TAO (3.7\%) to P-TAO (7.4\%) yields substantial gains (+10.9\% in novel AssocA), while further increasing to full C-TAO (100\%) provides additional improvements (+2.8\% in novel AssocA, for a total gain of +13.7\% from TAO to C-TAO).
This diminishing return pattern suggests that OVMOT training benefits significantly from \textit{any} continuous supervision, while fully-continuous annotations provide additional gains.
Even minimal pairwise continuous labels can effectively bridge the temporal gaps in sparse annotations, enabling models to learn basic motion dynamics.

Our method shows a similar trend but with more consistent improvements across all density levels, demonstrating better data efficiency.
The comparison also highlights that C-TAO's value extends beyond our approach: OVTrack improves from 25.4\% to 30.5\% novel TETA (+5.1\%) when trained on C-TAO versus original TAO, confirming the dataset's general utility for OVMOT training.
This observation suggests that significant performance gains are achievable with moderate annotation costs, providing practical guidance for future dataset construction efforts.

\subsection{Zero-Shot Cross-Domain Generalization}
\label{sec:BDD}
To evaluate the cross-domain generalization capability of our method, we conduct zero-shot experiments on the BDD100K dataset~\cite{yu2020bdd100k}, a large-scale driving video dataset with diverse scenarios.
Following the evaluation protocol in~\cite{li2023ovtrack}, we directly apply models trained on TAO (without BDD100K subset) to the BDD100K validation set without any fine-tuning or domain adaptation, evaluating their ability to generalize to the autonomous driving domain. 

\begin{table}[!t]
	\renewcommand{\arraystretch}{1.1}
	\footnotesize
	\centering
	\caption{Zero-shot cross-domain generalization results on BDD100K validation set. (\%)}
	\label{tab:bdd_zeroshot}
	\begin{tabular}{l|cccc}
		\hline
		\textbf{Method} & \textbf{TETA} & \textbf{LocA} & \textbf{AssocA} & \textbf{ClsA} \\
		\hline
		QDTrack~\cite{fischer2023qdtrack} & 32.0 & 25.9 & 27.8 & 42.4 \\
		TETer~\cite{li2022tracking} & 33.2 & 24.5 & 31.8 & 43.4 \\
		OVTrack~\cite{li2023ovtrack} & 42.5 & 41.0 & 36.7 & 49.7 \\
		OVTR~\cite{li2025ovtr} & 43.1 & 42.0 & 37.1 & 50.1 \\
		COVTrack~\cite{qian2025covtrack} & 45.8 & 44.1 & 39.4 & \textbf{53.9} \\
		Ours & \textbf{46.7} & \textbf{45.7} & \textbf{40.7} & {53.8} \\
		\hline
	\end{tabular}
\end{table}

As shown in Table~\ref{tab:bdd_zeroshot}, our method achieves superior zero-shot performance on BDD100K, significantly outperforming existing OVMOT methods.
Compared to OVTrack (42.5\%), our method achieves +4.2\% improvement in TETA, demonstrating strong cross-domain generalization.
More importantly, our extended framework (COVTrack++) further improves upon the conference version by +0.9\% TETA (45.8\% $\to$ 46.7\%), with notable gains in LocA (+1.6\%) and AssocA (+1.3\%).
These improvements validate that MGA and TCP not only enhance performance on the TAO benchmark but also generalize effectively to different domains.
The robust performance on driving scenarios, which exhibit distinct characteristics from TAO (e.g., faster motion, different object distributions, camera perspectives), demonstrates that our method learns domain-invariant tracking capabilities rather than overfitting to specific dataset characteristics.

\subsection{Qualitative Analysis}
\label{sec:vis}
\begin{figure}[H]
	\centering
	\includegraphics[width=\columnwidth]{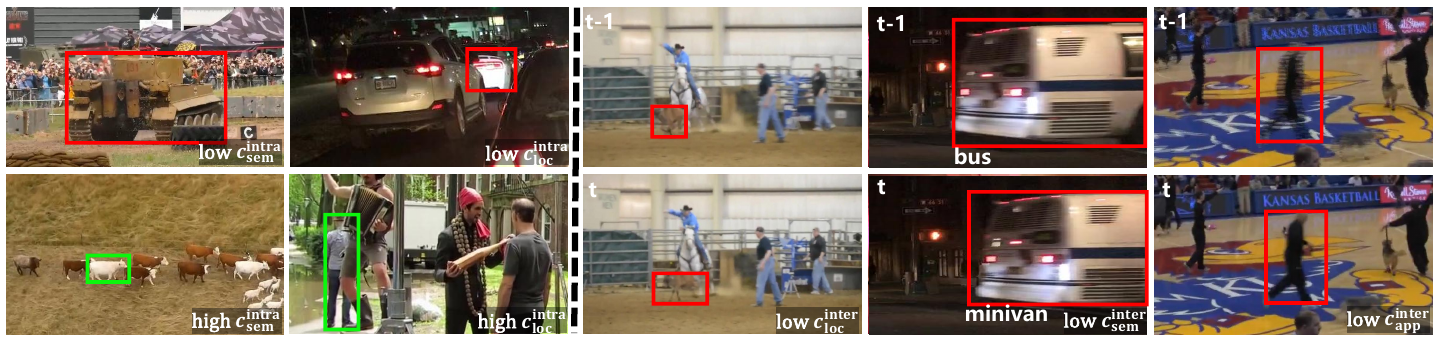}
	\caption{Visualization of intra-frame and inter-frame confidence in MCF. Left: intra-frame confidence. Right: inter-frame confidence.}
	\label{fig:conf_vis_intra_inter}
\end{figure}

\subsubsection{Intra-Frame Confidence Analysis}
Figure~\ref{fig:conf_vis_intra_inter} (left) illustrates the visualization of intra-frame confidence.
For semantic cues, low-confidence objects typically have ambiguous categories in complex backgrounds, while high-confidence objects are those with discriminative categories and simple backgrounds.
Regarding location cues, low confidence is often associated with unclear positions. The last example in Fig.~\ref{fig:conf_vis_intra_inter} (left) shows a high-confidence one, which has an accurate location while the appearance is disturbed. This way, the location cue can assist the tracking under this case.

\subsubsection{Inter-Frame Confidence Analysis}
Figure~\ref{fig:conf_vis_intra_inter} (right) shows several cases with low inter-frame confidence. Low $c^{\text{inter}}_{\text{loc}}$ values typically occur with rapid object motion causing significant inter-frame position variations. Low $c^{\text{inter}}_{\text{sem}}$ values appear when object categories are inconsistent across consecutive frames.
Low $c^{\text{inter}}_{\text{app}}$ values predominantly appear during sudden appearance changes, such as defocus or motion blur. 

\subsubsection{Hierarchical Multi-Granularity Aggregation Analysis}
\begin{figure}[!h]
	\centering
	\includegraphics[width=0.95\columnwidth]{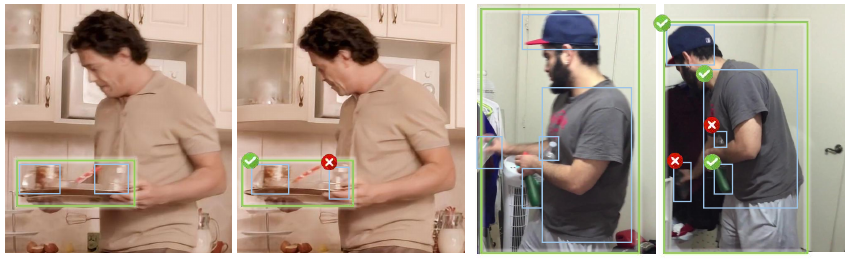}
	\caption{MGA visualization with two adjacent frames per case. Green: parent nodes. Blue: child nodes.}
	\label{fig:MGA_vis}
\end{figure}

We visualize the Multi-granularity aggregation (MGA) process in Fig.~\ref{fig:MGA_vis}. Each case shows two adjacent frames, where green boxes denote parent nodes and blue boxes denote child nodes. In the left case, a plate carries two cups; in the second frame, TAQ filters the partially occluded right cup and preserves the clearer left cup, enabling reliable aggregation for the plate and correcting the association. This scenario is challenging because the plate rotates and its appearance shifts with the objects it carries. In the right case, a person is placing a cup while under defocus; direct tracking of the person fails, while MGA aggregates distinctive child cues (hat, shirt, and green cup) and suppresses unreliable children (occluded bottle and blurry cup), enabling robust tracking in a challenging scene.

\subsubsection{Temporal Confidence Propagation Analysis}
\begin{figure}[!h]
	\centering
	\includegraphics[width=0.95\columnwidth]{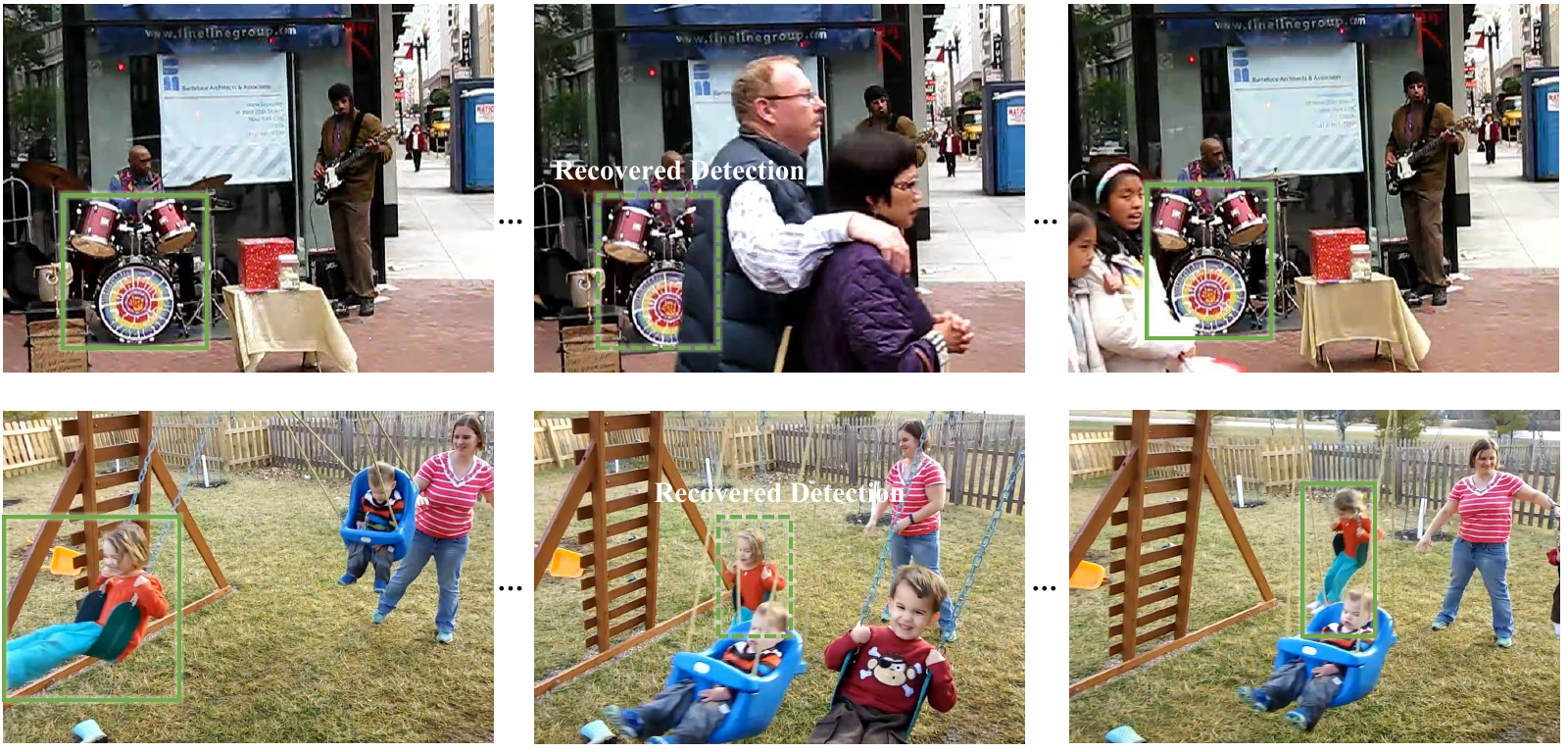}
	\caption{Visualization of TCP recovering the missed detection boxes (middle frame omitted).}
	\label{fig:tcp_vis}
\end{figure}

We visualize TCP in Fig.~\ref{fig:tcp_vis}. Each case shows three frames (with the middle frame omitted). In the top sequence, a drum is missed under occlusion, but TCP propagates confidence from temporally consistent cues to recover the detection and restore a complete trajectory; the drum's child parts provide robust cues that support the recovery. In the bottom sequence, a child on a swing is occluded by another child and the detection disappears; TCP recovers the track by leveraging reliable cues such as the girl's orange sweater, maintaining continuity even when the upper body is fully occluded.

\subsubsection{Real-World Visualization Results}
\begin{figure}[!h]
	\centering
	\includegraphics[width=0.95\columnwidth]{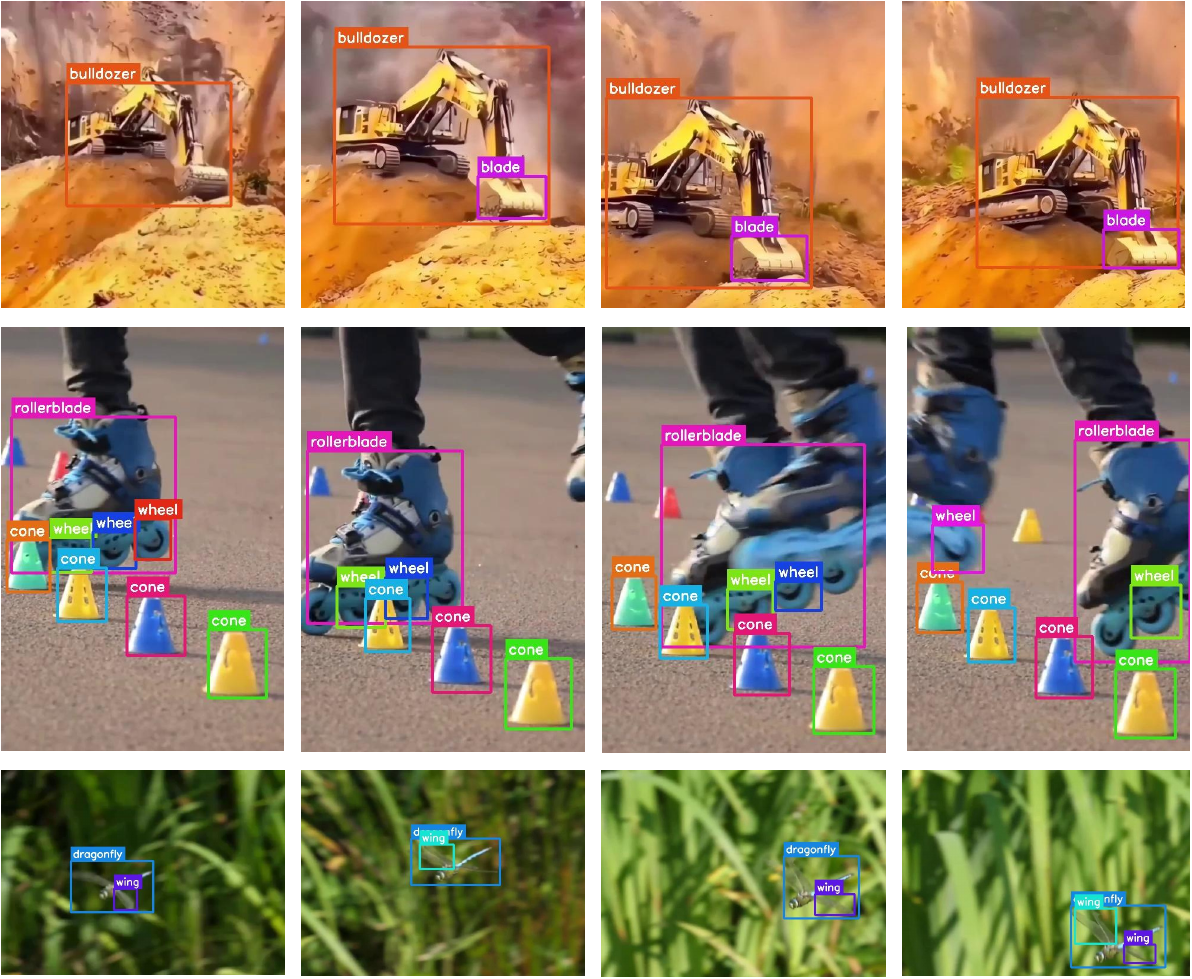}
	\caption{Visualization results of COVTrack++ on real-world videos.}
	\label{fig:vis_exam}
\end{figure}

Figure~\ref{fig:vis_exam} illustrates challenging tracking results involving novel objects under rapid motion and severe occlusion. By integrating the multi-cue association features and leveraging the hierarchical structure of dense detection, our method significantly enhances the tracking robustness. Notably, the proposed TCP generates superior object detection candidates for association, enabling the effective utilization of visible components to mitigate these disturbances.

\begin{figure*}[!t]
	\centering
	\includegraphics[width=0.86\textwidth]{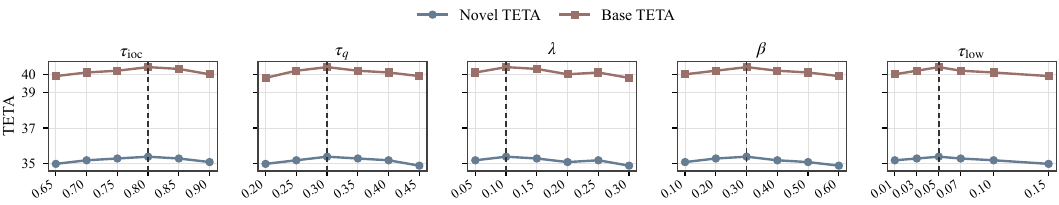}
	\caption{\revised{One-at-a-time hyperparameter sensitivity on TAO validation. The dashed line marks the selected value.}}
	\label{fig:hyper_sensitivity}
\end{figure*}

\subsection{\revised{Inference Speed and Hyperparameter Sensitivity}}
\label{sec:speed_hyper} 

\revised{Table~\ref{tab:speed} reports end-to-end inference speed on TAO validation videos under the same input size \(3\times800\times1334\) and single RTX 3090 GPU. COVTrack++ achieves 13.9 FPS, close to COVTrack (14.3 FPS) and within the fastest group among the compared methods.}   

\begin{table}[!h]
	\renewcommand{\arraystretch}{1.02}
	\scriptsize
	\centering
	\caption{\revised{Inference speed comparison on TAO validation set.}}
	\label{tab:speed}
	\setlength{\tabcolsep}{2.4pt}
	\begin{tabular}{l|cccccc}
		\hline
		\revised{\textbf{Method}} & \revised{QDTrack} & \revised{MASA} & \revised{OVTrack} & \revised{OVTR} & \revised{COVTrack} & \revised{COVTrack++} \\
		\hline
		\revised{\textbf{FPS}} & \revised{13.8} & \revised{13.4} & \revised{1.8} & \revised{3.4} & \revised{14.3} & \revised{13.9} \\
		\hline
	\end{tabular}
\end{table}

\revised{For hyperparameters, detector/NMS settings and other protocol constants follow the established detector/OVTrack settings. We tune only the method-specific parameters in Table~\ref{tab:hyper_protocol} on TAO validation, using novel TETA as the primary objective and base TETA as a stability check. Once selected, the same values are fixed for TAO test, BDD100K, dense C-TAO-val, and ablations. No TAO-test, BDD100K, or dense C-TAO-val metrics are used for selection, so no test leakage occurs.} 

\begin{table}[h]
	\renewcommand{\arraystretch}{1.05}
	\scriptsize
	\centering
	\caption{\revised{Search spaces and selected values of hyperparameters.}}
	\label{tab:hyper_protocol}
	\begin{tabular}{l|l|c}
		\hline
		\revised{\textbf{Parameter}} & \revised{\textbf{Search space}} & \revised{\textbf{Selected}} \\
		\hline
		\revised{\(\tau_{\mathrm{ioc}}\)} & \revised{\(\{0.65,0.70,0.75,0.80,0.85,0.90\}\)} & \revised{0.80} \\
		\revised{\(\tau_q\)} & \revised{\(\{0.20,0.25,0.30,0.35,0.40,0.45\}\)} & \revised{0.30} \\
		\revised{\(\lambda\)} & \revised{\(\{0.05,0.10,0.15,0.20,0.25,0.30\}\)} & \revised{0.10} \\
		\revised{\(\beta\)} & \revised{\(\{0.10,0.20,0.30,0.40,0.50,0.60\}\)} & \revised{0.30} \\
		\revised{\(\tau_{\mathrm{low}}\)} & \revised{\(\{0.01,0.03,0.05,0.07,0.10,0.15\}\)} & \revised{0.05} \\
		\hline
	\end{tabular}
\end{table}

\revised{As shown in Fig.~\ref{fig:hyper_sensitivity}, the selected values lie in stable high-performing regions, indicating that the method is not sensitive to narrow hyperparameter choices.} 

\newpage

\section{Conclusion}
In this work, we have addressed the fundamental challenges in open-vocabulary multi-object tracking through two key contributions. 
First, we constructed C-TAO, the first continuously annotated training dataset for OVMOT, which increased the annotation density by 26$\times$ over the original TAO and enabled effective learning of motion dynamics alongside diverse category information.
\revised{We further add C-TAO-val as a dense validation benchmark for temporal-continuity evaluation.}
Importantly, C-TAO has benefited not only our approach but also existing methods, confirming its general utility for the OVMOT community.
Second, we proposed COVTrack++, a unified and synergistic framework to {fully explore and leverage the coupling and dependence among different sub-tasks} in OVMOT.
This framework is achieved by the multi-cue adaptive fusion and multi-granularity hierarchical aggregation for association learning, and temporal confidence propagation for detection refinement.
Extensive experiments have demonstrated state-of-the-art performance on TAO (35.4\% novel TETA on validation, 30.5\% on test) and strong zero-shot generalization to BDD100K (46.7\% TETA), with substantial improvements over previous methods.


\bibliographystyle{IEEEtran}
\bibliography{main}

@String(CVPR= {IEEE Conf. Comput. Vis. Pattern Recog.})

@String(ECCV= {Eur. Conf. Comput. Vis.})

@String(ICIP = {IEEE Int. Conf. Image Process.})

@String(ICLR = {Int. Conf. Learn. Represent.})

@String(CVPR  = {CVPR})

@String(ECCV  = {ECCV})

@String(ICIP  = {ICIP})

@String(ICLR  = {ICLR})

@inproceedings{bergmann2019tracking,
  title={Tracking without bells and whistles},
  author={Bergmann, Philipp and Meinhardt, Tim and Leal-Taixe, Laura},
  booktitle={Proceedings of the IEEE/CVF international conference on computer vision},
  pages={941--951},
  year={2019}
}

@inproceedings{leal2016learning,
  title={Learning by tracking: Siamese CNN for robust target association},
  author={Leal-Taix{\'e}, Laura and Canton-Ferrer, Cristian and Schindler, Konrad},
  booktitle={Proceedings of the IEEE conference on computer vision and pattern recognition workshops},
  pages={33--40},
  year={2016}
}

@inproceedings{pang2021quasi,
  title={Quasi-dense similarity learning for multiple object tracking},
  author={Pang, Jiangmiao and Qiu, Linlu and Li, Xia and Chen, Haofeng and Li, Qi and Darrell, Trevor and Yu, Fisher},
  booktitle={Proceedings of the IEEE/CVF conference on computer vision and pattern recognition},
  pages={164--173},
  year={2021}
}

@inproceedings{sadeghian2017tracking,
  title={Tracking the untrackable: Learning to track multiple cues with long-term dependencies},
  author={Sadeghian, Amir and Alahi, Alexandre and Savarese, Silvio},
  booktitle={Proceedings of the IEEE international conference on computer vision},
  pages={300--311},
  year={2017}
}

@inproceedings{wojke2017simple,
  title={Simple online and realtime tracking with a deep association metric},
  author={Wojke, Nicolai and Bewley, Alex and Paulus, Dietrich},
  booktitle={2017 IEEE international conference on image processing (ICIP)},
  pages={3645--3649},
  year={2017},
  organization={IEEE}
}

@inproceedings{bewley2016simple,
  title={Simple online and realtime tracking},
  author={Bewley, Alex and Ge, Zongyuan and Ott, Lionel and Ramos, Fabio and Upcroft, Ben},
  booktitle={2016 IEEE international conference on image processing (ICIP)},
  pages={3464--3468},
  year={2016},
  organization={IEEE}
}

@inproceedings{xiao2018simple,
  title={Simple baselines for human pose estimation and tracking},
  author={Xiao, Bin and Wu, Haiping and Wei, Yichen},
  booktitle={Proceedings of the European conference on computer vision (ECCV)},
  pages={466--481},
  year={2018}
}

@inproceedings{li2024slack,
  title={SLAck: Semantic, Location, and Appearance Aware Open-Vocabulary Tracking},
  author={Li, Siyuan and Ke, Lei and Yang, Yung-Hsu and Piccinelli, Luigi and Seg{\`u}, Mattia and Danelljan, Martin and Van Gool, Luc},
  booktitle={Proceedings of the European conference on computer vision (ECCV)},
  year={2024}
}

@article{li2025open,
  title={Open-Vocabulary Multi-Object Tracking with Domain Generalized and Temporally Adaptive Features},
  author={Li, Run and Zhang, Dawei and Wang, Yanchao and Jiang, Yunliang and Zheng, Zhonglong and Jeon, Sang-Woon and Wang, Hua},
  journal={IEEE Transactions on Multimedia},
  year={2025},
  publisher={IEEE}
}

@inproceedings{li2025ovtr,
  title={OVTR: End-to-End Open-Vocabulary Multiple Object Tracking with Transformer},
  author={Li, Jinyang and Yu, En and Chen, Sijia and Tao, Wenbing},
  booktitle={The Thirteenth International Conference on Learning Representations (ICLR)},
  year={2025},
  note={arXiv preprint arXiv:2503.10616}
}

@inproceedings{li2025attention,
  title={Attention to Trajectory: Trajectory-Aware Open-Vocabulary Tracking},
  author={Li, Yunhao and Jiao, Yifan and Meng, Dan and Fan, Heng and Zhang, Libo},
  booktitle={Proceedings of the IEEE/CVF International Conference on Computer Vision},
  year={2025}
}

@article{luiten2020track,
  title={Track to reconstruct and reconstruct to track},
  author={Luiten, Jonathon and Fischer, Tobias and Leibe, Bastian},
  journal={IEEE Robotics and Automation Letters},
  volume={5},
  number={2},
  pages={1803--1810},
  year={2020},
  publisher={IEEE}
}

@inproceedings{meinhardt2022trackformer,
  title={Trackformer: Multi-object tracking with transformers},
  author={Meinhardt, Tim and Kirillov, Alexander and Leal-Taixe, Laura and Feichtenhofer, Christoph},
  booktitle={Proceedings of the IEEE/CVF conference on computer vision and pattern recognition},
  pages={8844--8854},
  year={2022}
}

@article{sun2020transtrack,
  title={Transtrack: Multiple object tracking with transformer},
  author={Sun, Peize and Cao, Jinkun and Jiang, Yi and Zhang, Rufeng and Xie, Enze and Yuan, Zehuan and Wang, Changhu and Luo, Ping},
  journal={arXiv preprint arXiv:2012.15460},
  year={2020}
}

@inproceedings{zeng2022motr,
  title={Motr: End-to-end multiple-object tracking with transformer},
  author={Zeng, Fangao and Dong, Bin and Zhang, Yuang and Wang, Tiancai and Zhang, Xiangyu and Wei, Yichen},
  booktitle={European Conference on Computer Vision},
  pages={659--675},
  year={2022},
  organization={Springer}
}

@inproceedings{zhou2022global,
  title={Global tracking transformers},
  author={Zhou, Xingyi and Yin, Tianwei and Koltun, Vladlen and Kr{\"a}henb{\"u}hl, Philipp},
  booktitle={Proceedings of the IEEE/CVF Conference on Computer Vision and Pattern Recognition},
  pages={8771--8780},
  year={2022}
}

@inproceedings{dave2020tao,
  title={Tao: A large-scale benchmark for tracking any object},
  author={Dave, Achal and Khurana, Tarasha and Tokmakov, Pavel and Schmid, Cordelia and Ramanan, Deva},
  booktitle={Computer Vision--ECCV 2020: 16th European Conference, Glasgow, UK, August 23--28, 2020, Proceedings, Part V 16},
  pages={436--454},
  year={2020},
  organization={Springer}
}

@article{du20211st,
  title={1st place solution to eccv-tao-2020: Detect and represent any object for tracking},
  author={Du, Fei and Xu, Bo and Tang, Jiasheng and Zhang, Yuqi and Wang, Fan and Li, Hao},
  journal={arXiv preprint arXiv:2101.08040},
  year={2021}
}

@inproceedings{li2022tracking,
  title={Tracking every thing in the wild},
  author={Li, Siyuan and Danelljan, Martin and Ding, Henghui and Huang, Thomas E and Yu, Fisher},
  booktitle={European Conference on Computer Vision},
  pages={498--515},
  year={2022},
  organization={Springer}
}

@inproceedings{radford2021learning,
  title={Learning transferable visual models from natural language supervision},
  author={Radford, Alec and Kim, Jong Wook and Hallacy, Chris and Ramesh, Aditya and Goh, Gabriel and Agarwal, Sandhini and Sastry, Girish and Askell, Amanda and Mishkin, Pamela and Clark, Jack and others},
  booktitle={International conference on machine learning},
  pages={8748--8763},
  year={2021},
  organization={PMLR}
}

@inproceedings{
gu2021open,
title={Open-vocabulary Object Detection via Vision and Language Knowledge Distillation},
author={Xiuye Gu and Tsung-Yi Lin and Weicheng Kuo and Yin Cui},
booktitle={International Conference on Learning Representations},
year={2022}
}

@inproceedings{mitzel2012taking,
  title={Taking mobile multi-object tracking to the next level: People, unknown objects, and carried items},
  author={Mitzel, Dennis and Leibe, Bastian},
  booktitle={Computer Vision--ECCV 2012: 12th European Conference on Computer Vision, Florence, Italy, October 7-13, 2012, Proceedings, Part V 12},
  pages={566--579},
  year={2012},
  organization={Springer}
}

@inproceedings{ovsep2018track,
  title={Track, then decide: Category-agnostic vision-based multi-object tracking},
  author={O{\v{s}}ep, Aljo{\v{s}}a and Mehner, Wolfgang and Voigtlaender, Paul and Leibe, Bastian},
  booktitle={2018 IEEE International Conference on Robotics and Automation (ICRA)},
  pages={3494--3501},
  year={2018},
  organization={IEEE}
}

@inproceedings{liu2022opening,
  title={Opening up open world tracking},
  author={Liu, Yang and Zulfikar, Idil Esen and Luiten, Jonathon and Dave, Achal and Ramanan, Deva and Leibe, Bastian and O{\v{s}}ep, Aljo{\v{s}}a and Leal-Taix{\'e}, Laura},
  booktitle={Proceedings of the IEEE/CVF Conference on Computer Vision and Pattern Recognition},
  pages={19045--19055},
  year={2022}
}

@article{fischer2023qdtrack,
  title={Qdtrack: Quasi-dense similarity learning for appearance-only multiple object tracking},
  author={Fischer, Tobias and Huang, Thomas E and Pang, Jiangmiao and Qiu, Linlu and Chen, Haofeng and Darrell, Trevor and Yu, Fisher},
  journal={IEEE Transactions on Pattern Analysis and Machine Intelligence},
  year={2023},
  publisher={IEEE}
}

@article{zhang2021fairmot,
  title={Fairmot: On the fairness of detection and re-identification in multiple object tracking},
  author={Zhang, Yifu and Wang, Chunyu and Wang, Xinggang and Zeng, Wenjun and Liu, Wenyu},
  journal={International Journal of Computer Vision},
  volume={129},
  pages={3069--3087},
  year={2021},
  publisher={Springer}
}

@inproceedings{kirillov2023segment,
  title={Segment anything},
  author={Kirillov, Alexander and Mintun, Eric and Ravi, Nikhila and Mao, Hanzi and Rolland, Chloe and Gustafson, Laura and Xiao, Tete and Whitehead, Spencer and Berg, Alexander C and Lo, Wan-Yen and others},
  booktitle={Proceedings of the IEEE/CVF international conference on computer vision},
  pages={4015--4026},
  year={2023}
}

@inproceedings{cai2022memot,
  title={Memot: Multi-object tracking with memory},
  author={Cai, Jiarui and Xu, Mingze and Li, Wei and Xiong, Yuanjun and Xia, Wei and Tu, Zhuowen and Soatto, Stefano},
  booktitle={Proceedings of the IEEE/CVF Conference on Computer Vision and Pattern Recognition},
  pages={8090--8100},
  year={2022}
}

@inproceedings{qin2023motiontrack,
  title={Motiontrack: Learning robust short-term and long-term motions for multi-object tracking},
  author={Qin, Zheng and Zhou, Sanping and Wang, Le and Duan, Jinghai and Hua, Gang and Tang, Wei},
  booktitle={Proceedings of the IEEE/CVF conference on computer vision and pattern recognition},
  pages={17939--17948},
  year={2023}
}

@article{du2023strongsort,
  title={Strongsort: Make deepsort great again},
  author={Du, Yunhao and Zhao, Zhicheng and Song, Yang and Zhao, Yanyun and Su, Fei and Gong, Tao and Meng, Hongying},
  journal={IEEE Transactions on Multimedia},
  year={2023},
  publisher={IEEE}
}

@inproceedings{zhou2020tracking,
  title={Tracking objects as points},
  author={Zhou, Xingyi and Koltun, Vladlen and Kr{\"a}henb{\"u}hl, Philipp},
  booktitle={European conference on computer vision},
  pages={474--490},
  year={2020},
  organization={Springer}
}

@inproceedings{saleh2021probabilistic,
  title={Probabilistic tracklet scoring and inpainting for multiple object tracking},
  author={Saleh, Fatemeh and Aliakbarian, Sadegh and Rezatofighi, Hamid and Salzmann, Mathieu and Gould, Stephen},
  booktitle={Proceedings of the IEEE/CVF conference on computer vision and pattern recognition},
  pages={14329--14339},
  year={2021}
}

@article{wang2023camo,
  title={Camo-mot: Combined appearance-motion optimization for 3d multi-object tracking with camera-lidar fusion},
  author={Wang, Li and Zhang, Xinyu and Qin, Wenyuan and Li, Xiaoyu and Gao, Jinghan and Yang, Lei and Li, Zhiwei and Li, Jun and Zhu, Lei and Wang, Hong and others},
  journal={IEEE Transactions on Intelligent Transportation Systems},
  year={2023},
  publisher={IEEE}
}

@inproceedings{huang2023delving,
  title={Delving into Motion-Aware Matching for Monocular 3D Object Tracking},
  author={Huang, Kuan-Chih and Yang, Ming-Hsuan and Tsai, Yi-Hsuan},
  booktitle={Proceedings of the IEEE/CVF International Conference on Computer Vision},
  pages={6909--6918},
  year={2023}
}

@article{krejvci2024pedestrian,
  title={Pedestrian Tracking with Monocular Camera using Unconstrained 3D Motion Model},
  author={Krej{\v{c}}{\'\i}, Jan and Kost, Oliver and Straka, Ond{\v{r}}ej and Dun{\'\i}k, Jind{\v{r}}ich},
  journal={arXiv preprint arXiv:2403.11978},
  year={2024}
}

@inproceedings{andriluka2008people,
  title={People-tracking-by-detection and people-detection-by-tracking},
  author={Andriluka, Mykhaylo and Roth, Stefan and Schiele, Bernt},
  booktitle={2008 IEEE Conference on computer vision and pattern recognition},
  pages={1--8},
  year={2008},
  organization={IEEE}
}

@inproceedings{li2023ovtrack,
  title={Ovtrack: Open-vocabulary multiple object tracking},
  author={Li, Siyuan and Fischer, Tobias and Ke, Lei and Ding, Henghui and Danelljan, Martin and Yu, Fisher},
  booktitle={Proceedings of the IEEE/CVF conference on computer vision and pattern recognition},
  pages={5567--5577},
  year={2023}
}

@inproceedings{he2016deep,
  title={Deep residual learning for image recognition},
  author={He, Kaiming and Zhang, Xiangyu and Ren, Shaoqing and Sun, Jian},
  booktitle={Proceedings of the IEEE conference on computer vision and pattern recognition},
  pages={770--778},
  year={2016}
}

@inproceedings{gupta2019lvis,
  title={Lvis: A dataset for large vocabulary instance segmentation},
  author={Gupta, Agrim and Dollar, Piotr and Girshick, Ross},
  booktitle={Proceedings of the IEEE/CVF conference on computer vision and pattern recognition},
  pages={5356--5364},
  year={2019}
}

@article{dendorfer2020mot20,
  title={Mot20: A benchmark for multi object tracking in crowded scenes},
  author={Dendorfer, Patrick and Rezatofighi, Hamid and Milan, Anton and Shi, Javen and Cremers, Daniel and Reid, Ian and Roth, Stefan and Schindler, Konrad and Leal-Taix{\'e}, Laura},
  journal={arXiv preprint arXiv:2003.09003},
  year={2020}
}

@inproceedings{sun2022dancetrack,
  title={Dancetrack: Multi-object tracking in uniform appearance and diverse motion},
  author={Sun, Peize and Cao, Jinkun and Jiang, Yi and Yuan, Zehuan and Bai, Song and Kitani, Kris and Luo, Ping},
  booktitle={Proceedings of the IEEE/CVF Conference on Computer Vision and Pattern Recognition},
  pages={20993--21002},
  year={2022}
}

@inproceedings{segu2024walker,
  title={Walker: self-supervised multiple object tracking by walking on temporal appearance graphs},
  author={Segu, Mattia and Piccinelli, Luigi and Li, Siyuan and Van Gool, Luc and Yu, Fisher and Schiele, Bernt},
  booktitle={European Conference on Computer Vision},
  pages={1--18},
  year={2024},
  organization={Springer}
}

@inproceedings{wang2020towards,
  title={Towards real-time multi-object tracking},
  author={Wang, Zhongdao and Zheng, Liang and Liu, Yixuan and Li, Yali and Wang, Shengjin},
  booktitle={European conference on computer vision},
  pages={107--122},
  year={2020},
  organization={Springer}
}

@inproceedings{zhang2022bytetrack,
  title={Bytetrack: Multi-object tracking by associating every detection box},
  author={Zhang, Yifu and Sun, Peize and Jiang, Yi and Yu, Dongdong and Weng, Fucheng and Yuan, Zehuan and Luo, Ping and Liu, Wenyu and Wang, Xinggang},
  booktitle={European conference on computer vision},
  pages={1--21},
  year={2022},
  organization={Springer}
}

@inproceedings{Li_2024_CVPR,
  title={Matching Anything by Segmenting Anything},
  author={Li, Siyuan and Ke, Lei and Danelljan, Martin and Piccinelli, Luigi and Segu, Mattia and Van Gool, Luc and Yu, Fisher},
  booktitle={Proceedings of the IEEE/CVF Conference on Computer Vision and Pattern Recognition},
  pages={18963--18973},
  year={2024}
}

@inproceedings{cao2023observation,
  title={Observation-centric sort: Rethinking sort for robust multi-object tracking},
  author={Cao, Jinkun and Pang, Jiangmiao and Weng, Xinshuo and Khirodkar, Rawal and Kitani, Kris},
  booktitle={Proceedings of the IEEE/CVF conference on computer vision and pattern recognition},
  pages={9686--9696},
  year={2023}
}

@article{geiger2013vision,
  title={Vision meets robotics: The kitti dataset},
  author={Geiger, Andreas and Lenz, Philip and Stiller, Christoph and Urtasun, Raquel},
  journal={The international journal of robotics research},
  volume={32},
  number={11},
  pages={1231--1237},
  year={2013},
  publisher={Sage Publications Sage UK: London, England}
}

@inproceedings{qian2025covtrack,
  title={COVTrack: Continuous Open-Vocabulary Tracking via Adaptive Multi-Cue Fusion},
  author={Qian, Zekun and Han, Ruize and Wang, Zhixiang and Hou, Junhui and Feng, Wei},
  booktitle={Proceedings of the IEEE/CVF International Conference on Computer Vision},
  pages={10054--10063},
  year={2025}
}

@inproceedings{yu2020bdd100k,
  title={BDD100K: A Diverse Driving Dataset for Heterogeneous Multitask Learning},
  author={Yu, Fisher and Chen, Haofeng and Wang, Xin and Xian, Wenqi and Chen, Yingying and Liu, Fangchen and Madhavan, Vashisht and Darrell, Trevor},
  booktitle={IEEE/CVF Conference on Computer Vision and Pattern Recognition (CVPR)},
  pages={2636--2645},
  year={2020}
}

@inproceedings{wang2019learning,
  title={Learning Correspondence from the Cycle-Consistency of Time},
  author={Wang, Xiaolong and Jabri, Allan and Efros, Alexei A.},
  booktitle={Proceedings of the IEEE/CVF Conference on Computer Vision and Pattern Recognition},
  pages={2566--2576},
  year={2019}
}

\vspace{-0.5in}

\end{document}